\documentclass[10pt,twocolumn,letterpaper]{article}

\usepackage[pagenumbers]{iccv} 

\usepackage{multirow}
\usepackage[normalem]{ulem}
\usepackage{bm}

\definecolor{iccvblue}{rgb}{0.21,0.49,0.74}
\usepackage[pagebackref,breaklinks,colorlinks,allcolors=iccvblue]{hyperref}
\usepackage{marvosym}
\usepackage{hyperref}

\def\paperID{***} 
\def\confName{}
\def\confYear{2025}

\title{MVGSR: Multi-View Consistency Gaussian Splatting for Robust Surface Reconstruction }

\author{
Chenfeng Hou$^{1}$ \qquad Qi Xun Yeo$^{2}$ \qquad \qquad Mengqi Guo$^{2}$ \qquad \qquad Yongxin Su$^{1}$ \\[4pt]
{Yanyan Li}\textsuperscript{2,\Letter}  \qquad \qquad {Gim Hee Lee}$^{2}$ \\[4pt]
$^1$Beihang University \quad  $^2$National University of Singapore\\
\url{https://mvgsr.github.io}
}

\begin{document}
\twocolumn[{
   \maketitle
    \begin{center}
        \centering
        \vspace{-15pt} 
        \includegraphics[width=0.9\textwidth]{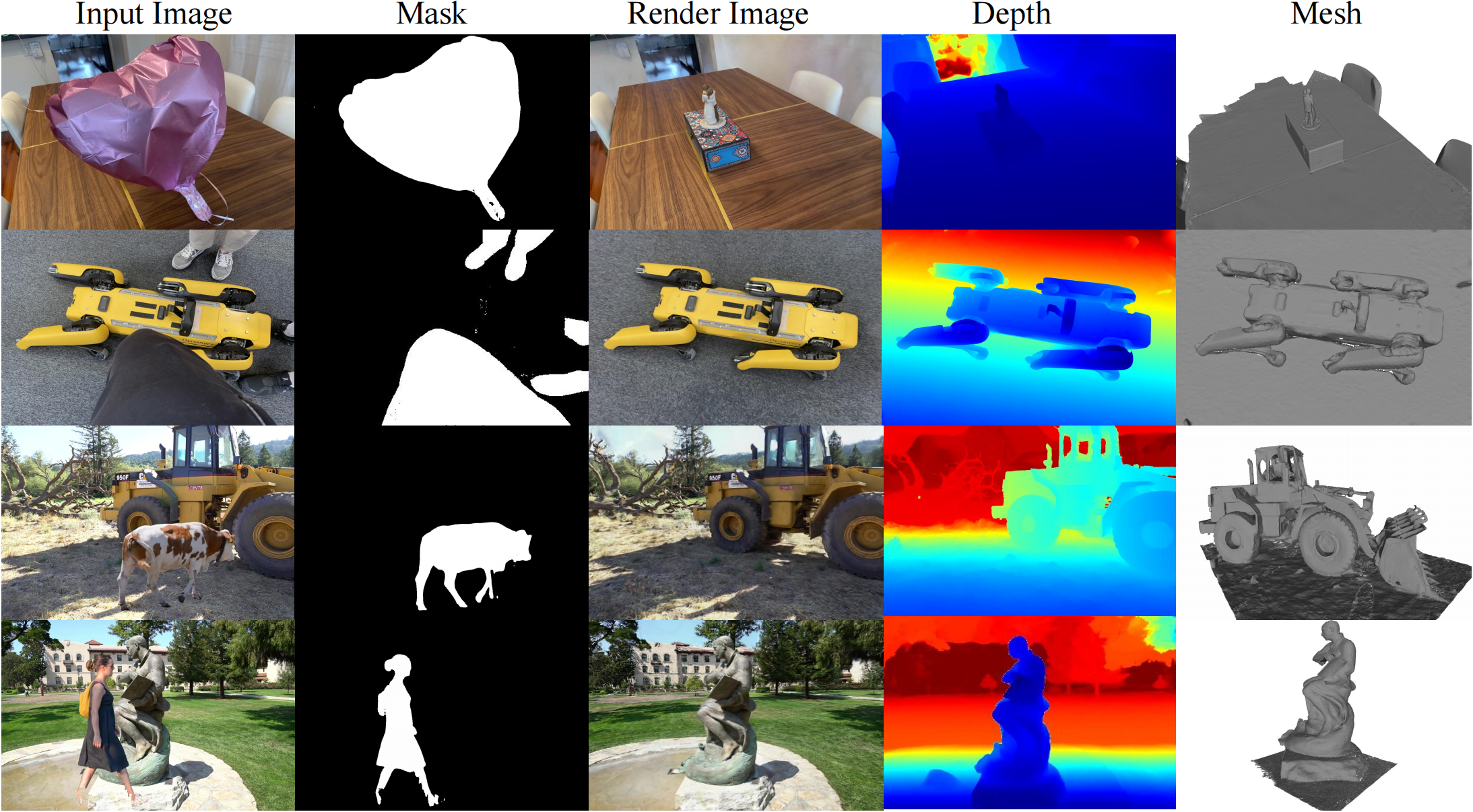}
        \captionof{figure}{MVGSR Performance Across Datasets: RobustNeRF\cite{sabour2023robustnerf} (line 1), On-the-Go\cite{Ren2024NeRFonthego} (line 2), and TnT~\cite{tnt} (lines 3-4). When input images contain distractors, MVGSR generates distractor masks through multi-view feature contrast, effectively preventing gradient leakage and thereby achieving high-quality surface reconstruction with photorealistic rendering fidelity}
        \label{fig:teaser}
    \end{center}
}]
\begin{abstract}
3D Gaussian Splatting (3DGS) has gained significant attention for its high-quality rendering capabilities, ultra-fast training, and inference speeds. 
However, when we apply 3DGS to surface reconstruction tasks, especially in environments with dynamic objects and distractors, the method suffers from floating artifacts and color errors due to inconsistency from different viewpoints.
To address this challenge, we propose Multi-View Consistency Gaussian Splatting for the domain of Robust Surface Reconstruction (\textbf{MVGSR}), which takes advantage of lightweight Gaussian models and a {heuristics-guided distractor masking} strategy for robust surface reconstruction in non-static environments.
Compared to existing methods that rely on MLPs for distractor segmentation strategies, our approach separates distractors from static scene elements by comparing multi-view feature consistency, allowing us to obtain precise distractor masks early in training. 
Furthermore, we introduce a pruning measure based on multi-view contributions to reset transmittance, effectively reducing floating artifacts. 
Finally, a multi-view consistency loss is applied to achieve high-quality performance in surface reconstruction tasks. 
Experimental results demonstrate that MVGSR achieves competitive geometric accuracy and rendering fidelity compared to the state-of-the-art surface reconstruction algorithms. 
More information is available on our project page (\href{https://mvgsr.github.io}{this url}).

\end{abstract}

\section{Introduction}
\label{sec:intro}
\begin{figure}
\centering
\includegraphics[width=\linewidth]{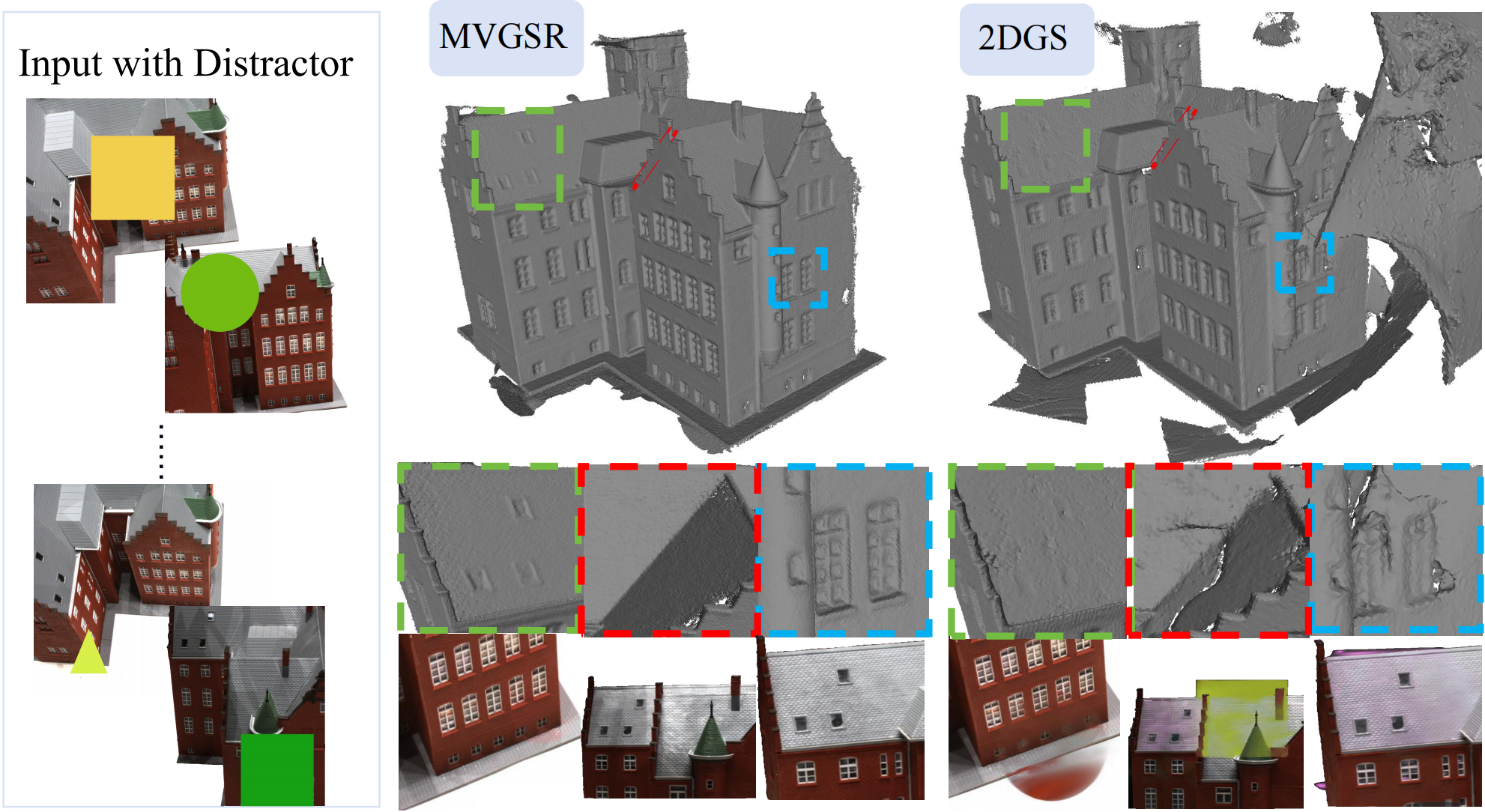}
\caption{Surface reconstruction in scenes with distractors. }
\label{fig:reconstruction-fig}
\end{figure}

In recent years, neural network-based implicit representations~\cite{mildenhall2021nerf,rosinol2023nerf,Di_2024_CVPR} of 3D scenes have gained significant popularity due to their impressive modeling capabilities and generalization power. More recently, 3D Gaussian-based representations~\cite{kerbl20233d,li2024geogaussian,huang20242d} have emerged as a promising alternative, offering a new approach to addressing this challenge. Compared to neural implicit representations, 3D Gaussian-based scene representations exhibit faster processing and higher rendering quality.

Although 3D Gaussian-based scene representations achieve impressive visual results, they have limitations~\cite{li2024geogaussian,cheng2024gaussianpro} in accurately describing a scene's geometry. This deficiency, particularly in the context of robust surface reconstruction, significantly impacts the overall quality of 3D reconstruction. In casually captured scenes~\cite{zhang2024streetgaussians3dobject,zhang2025gaussian}, moving pedestrians, vehicles, and other objects in the images are unavoidable distractors. These distractors occlude the central objects of interest. As shown in Fig.~\ref{fig:reconstruction-fig}, when there are viewpoint-dependent distractors, 3DGS-based surface reconstruction algorithms may model them as artifacts clustered in front of the camera lens or as viewpoint-dependent color representations attached to the object's surface. To address the issue of distractors during reconstruction, we propose a surface reconstruction algorithm that avoids these distractions, preventing artifacts while achieving competitive geometric accuracy and enhanced rendering capabilities.

Different to previous work~\cite{sabour2023robustnerf, sabourgoli2024spotlesssplats} using photometric residual decomposition to obtain distractor masks, our core insight is that {distractors appearing in only a few images lack semantic consistency across different views}, {resulting in noticeable differences in features extracted by pre-trained base models}. We leverage this observation to separate distractors from static scenes the early stage of training. Compared to iterative masks learned by multi-layer perceptrons (MLP) during training, masks obtained through multi-view comparison have two advantages: first, they distinguish distractors from high-frequency details, allowing continuous optimization of high-quality surface reconstruction. Second, strict mask boundaries prevent gradient leakage during Gaussian splitting, avoiding artifacts and viewpoint-dependent color errors in distractor regions.

To address the floating artifacts and ghosting caused by uncertain geometric relationships, we design a pruning measure based on multi-view contributions. We reset the transmittance of objects occluding the camera's view to mitigate the impact of occlusion on gradient flow. This strategy can remove floating artifacts while compressing the number of point clouds, with only a minimal decrease in rendering quality.

For high-precision optimization of surface reconstruction, we use a multi-view consistency loss function to enhance the reconstruction capability of Gaussian splats. This is achieved through structural and color consistency constraints on corresponding points from different views. In scenes with distractors, our method achieves optimal rendering quality and competitive reconstruction results. Our contributions can be summarized as follows:

\begin{itemize}

\item We propose a method using multi-view consistency to distinguish distractors from static objects before surface reconstruction, achieving strict distractor masking and significantly reducing floating artifacts and color errors. 

\item We introduce a new Gaussian pruning technique based on multi-view contributions to remove floating artifacts with minimal reduction in rendering capability.  

\item In scenes with distractors, we achieve high-fidelity rendering quality and high-precision reconstruction results. Our code and generated dataset will be released to the community.

\end{itemize}
\section{Related Work}
\label{sec:related}
\subsection{Traditional 3D Reconstruction }

Traditional Surface Reconstruction methods~\cite{ullman1979interpretation,schonberger2016structure,seitz2006comparison,yunus2021manhattanslam} can be roughly grouped into different classes based on their intermediate representations , such as point clouds~\cite{schonberger2016structure}, volumes~\cite{yunus2021manhattanslam}, and depth maps\cite{izadi2011kinectfusion}. These methods typically decompose the entire multi-view stereo (MVS) problem into several stages. First, a dense point cloud is extracted from the multi-view images via patch-based matching~\cite{mur2015probabilistic}. Then, the surface structure is constructed through either feature triangulation~\cite{ma2018sparse} or implicit surface fitting~\cite{ilic2005implicit}. These methods are often affected by mismatching or noise introduced during the reconstruction pipeline. 

\subsection{Neural Surface Reconstruction}

NeRF-based methods~\cite{mildenhall2021nerf,barron2022mip} take 5D rays as input to predict density and color sampled in implicit space, leading to more realistic rendering results. Despite NeRF's impressive performance in 3D reconstruction, its implicit function representation through volume rendering results in poor geometric accuracy and susceptibility to noise. To address these issues, methods such as NeuS~\cite{wang2021neus}, BakedSDF~\cite{yariv2023bakedsdf}, and UNISURF~\cite{oechsle2021unisurf} represent surfaces using signed distance functions (SDF) to achieve more accurate scene geometry. 
Meanwhile, Nerf2Mesh~\cite{tang2022nerf2mesh} introduces an iterative mesh refinement algorithm that adaptively adjusts vertex positions and volume density based on reprojection rendering error.However, while the NeRF-based framework exhibits powerful surface reconstruction capabilities, the stacked MLP layers impose limitations on inference time and representation power.

\begin{figure*}[]
  \centering
  \includegraphics[width=\textwidth]{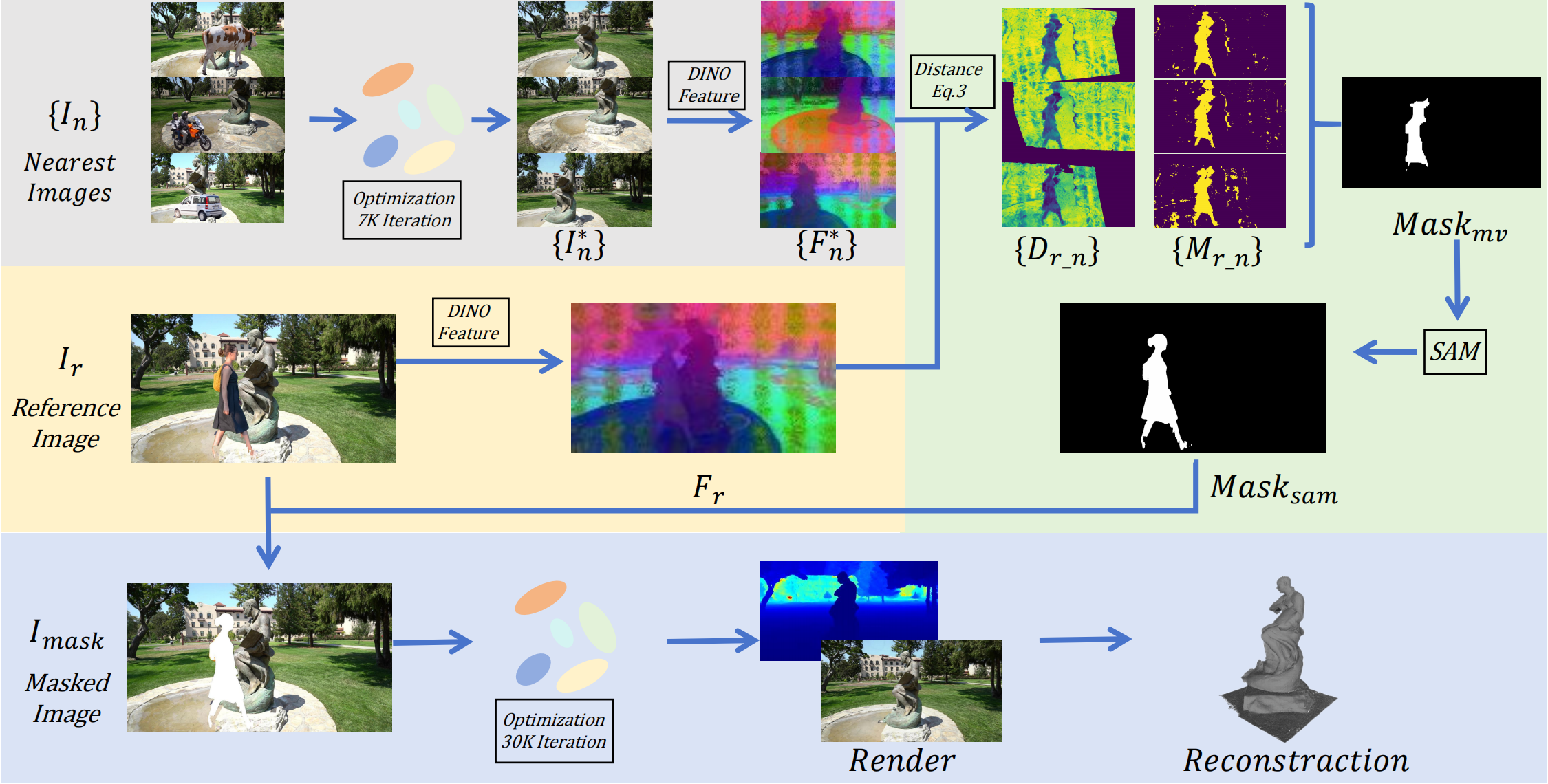} 
  \caption{The detailed architecture of our MVGSR Framework. Images with distractors are fed to the system that makes use of multi-view consistency Gaussian Splatting algorithm to achieve robust surface reconstruction for non-static environments.   }
  \label{fig:Dynamic Gaussian Splatting}
\end{figure*}

\subsection{GS-based Surface Reconstruction}
Different from neural surface reconstruction methods, GS-based approaches optimize point-based radiance fields, including 3D Gaussians, 2D Gaussians, and Gaussian surfels. SuGaR~\cite{guedon2024sugar} extracts meshes from the positions of 3D Gaussians and introduces a regularization term to encourage Gaussians to align with the scene surface based on sampled 3D point clouds. While this alignment improves geometric reconstruction accuracy, the irregular shapes of 3D Gaussians pose challenges in modeling smooth geometric surfaces.
2DGS~\cite{huang20242d} achieves view-consistent geometry by collapsing 3D volumes into a set of 2D Gaussian. GOF~\cite{yu2024gaussian} constructs Gaussian opacity fields and extracts 3D models directly from them. However, these 3DGS-based methods typically produce biased depth estimates and struggle with maintaining multi-view geometric consistency. To address the inconsistent issue in scene reconstruction, PGSR~\cite{chen2024pgsr} flattens the Gaussian function into a planar shape based on the assumption that scenes compose of several smaller planar regions. Leveraging these 3D relationships, PGSR introduces a new depth computation strategy to accurately extract geometric parameters from Gaussian surfels. 
Instead of assuming planar surfaces, our method proposes a more general architecture for object reconstruction, which incrementally transforms inconsistent ellipsoids to Gaussian surfels during the densification process. 

\subsection{Distractor Removal}
For reconstruction in non-static scenes, distractor removal is one of the most important task. There are two main strategies in dealing with this problem namely segmentation-based methods and heuristics-based methods. For the former, deep semantic or segmentation models are used to detect object with potential dynamic characteristics or to recognize static scenarios. Based on pre-trained models, DynaMoN~\cite{karaoglu2023dynamon} further utilized semantic maps to reconstruct dynamic scenes via Motion-Aware Fast and Robust Camera Localization for in dynamic NeRF. However, for the latter, approaches~\cite{martin2021nerf, sabour2023robustnerf, chen2024nerf} make use of heuristics generated from multi-view geometry algorithms to separate dynamic objects from static scenes. Specifically, NeRF-W~\cite{martin2021nerf} assumes that most of transient objects are generally small during the NeRF training process, while RobustNeRF~\cite{sabour2023robustnerf} tries to define transient objects
from static background based on the photometric residuals during the optimization module. Compared to RobustNeRF, NeRF-HUGS~\cite{chen2024nerf} leverages on masks estimated based pre-trained model with residual mask obtained from RobustNeRF to achieve more accurate distractor masking performance.

\section{Preliminary of Gaussian Splatting}
3DGS~\cite{kerbl20233d} models 3D scenes by employing a set of 3D Gaussians \(\{\mathcal{G}_i\}\). A Gaussian function defines each of these Gaussians at point \(\bm{p}_i \in \mathcal{P}\) :
$$
\mathcal G_i({\bm{x}}|\bm{\mu}_i, \bm{\Sigma_i}) = e^{-\frac{1}{2}(\bm{x} - \bm{\mu}_i)^\top\bm{\Sigma}_i^{-1}(\bm{x}-\bm{\mu}_i)},
$$
Each point \(\bm{p}_i \in \mathcal{P}\) is centered at \(\bm{\mu}_i \in \mathbb{R}^3\) with a corresponding 3D covariance matrix \(\bm{\Sigma}_i \in \mathbb{R}^{3 \times 3}\). This covariance matrix \(\bm{\Sigma}_i\) is factorized into a scaling matrix $\bm{S}_i\in \mathbb R^{3\times3}$ and a rotation matrix $\bm{R}_i\in \mathbb SO(3)$:
$$
\bm{\Sigma}_i = \bm{R}_i\bm{S}_i\bm{S}_i^\top \bm{R}_i^\top.
$$

3DGS facilitates quick \(\alpha\)-blending for rendering views. The transformation matrix \(W\) and intrinsic matrix \(\bm{K}\) allow \(\bm{\mu}_i\) and \(\bm{\Sigma}_i\) to be converted into camera coordinates related to \(W\), and subsequently projected into 2D coordinates using the following functions:
$$
\bm{\mu}_i^{'}=\bm{KW}[\bm{\mu}_i,1]^\top, \quad  \bm{\Sigma}_i^{'}=\bm{JW\Sigma}_i\bm{W}^\top \bm{J}^\top,
$$
where \(\bm{J}\) represents the Jacobian of the affine approximation for the projective transformation. The color \(\bm{C} \in \mathbb{R}^3\) of a pixel \(\bm{u}\) can be rendered using \(\alpha\)-blending.:
$$
\bm{C} = \sum_{i\in N} T_i \alpha_i \bm{c}_i,\quad T_i=\prod_{j=1}^{i-1}(1 - \alpha_i),
$$
Here, \(\alpha_i\) is determined by evaluating \(\mathcal{G}_i(\bm{u}|\bm{\mu}_i^{'}, \bm{\Sigma}_i^{'})\) and multiplying it with a learnable opacity associated with \(\mathcal{G}_i\). The view-dependent color \(\bm{c}_i\) is expressed using spherical harmonics (SH) from the Gaussian \(\mathcal{G}_i\). \(T_i\) represents the cumulative opacity, and \(N\) denotes the number of Gaussians the ray intersects.

The center $\bm{\mu}_i$ of a Gaussian $\mathcal G_i$ can be transformed into the camera coordinate system as:
$$
\begin{bmatrix}
    x_i,y_i,z_i,1
\end{bmatrix}^\top=\bm{[W|t]}[\bm{\mu}_i,1]^\top,
$$
And previous methods~\cite{jiang2023gaussianshader,cheng2024gaussianpro} in depth rendering under the current viewpoint can be denoted as:
$$
\bm{D} = \sum_{i\in N} T_i \alpha_i z_i.
\label{eq:render_depth}
$$
For 3D Gaussians, the direction of the minimum scale factor corresponds to the normal \( n_i \) of the Gaussian. The normal map under the current viewpoint is achieved through \(\alpha\)-blending:
$$
\boldsymbol{N}=\sum_{i \in N} T_i  \alpha_{i} \boldsymbol{R}_{c}^{T} \boldsymbol{n}_{i}
$$
where $R_c$ is the rotation from the camera to the global world.
\section{Methodology}

Given a set of images of a scene containing outliers, the goal of our method is to achieve high-precision surface reconstruction while maintaining robust view rendering performance. In existing work on outlier masking~\cite{sabour2023robustnerf,sabourgoli2024spotlesssplats}, mask estimation typically relies on photometric errors, but these methods cannot accurately distinguish between outliers and reconstruction details, leading to potential inaccuracies.
To address this issue, we propose a novel method that leverages feature similarity across multiple views to differentiate outliers. Specifically, we utilize features extracted from a self-supervised 2D foundation model~\cite{oquab2023dinov2} for feature extraction. With fewer training iterations, we can generate a rough scene representation to compute the mapping between the original view and adjacent views. By comparing feature similarities, we can derive an initial mask that helps identify outliers.
To refine the mask boundaries and enhance its accuracy, we sample points from the mask and apply a basic segmentation model to obtain precise mask results. Retraining with this refined mask helps mitigate gradient leakage during the optimization process. Additionally, to further reduce artifacts, we apply pruning based on multi-view contributions, ensuring the removal of unreliable information from the reconstruction.
The process of obtaining the initial mask will be detailed in Sec.~\ref{sec:method:distractor-detection}, while Sec.~\ref{sec:method:mulview-prune} will discuss multi-view pruning, and Sec.~\ref{sec:method:mulview-loss} will describe the loss function used in the optimization.

\begin{figure}
    \centering
    \includegraphics[width=0.8\linewidth]{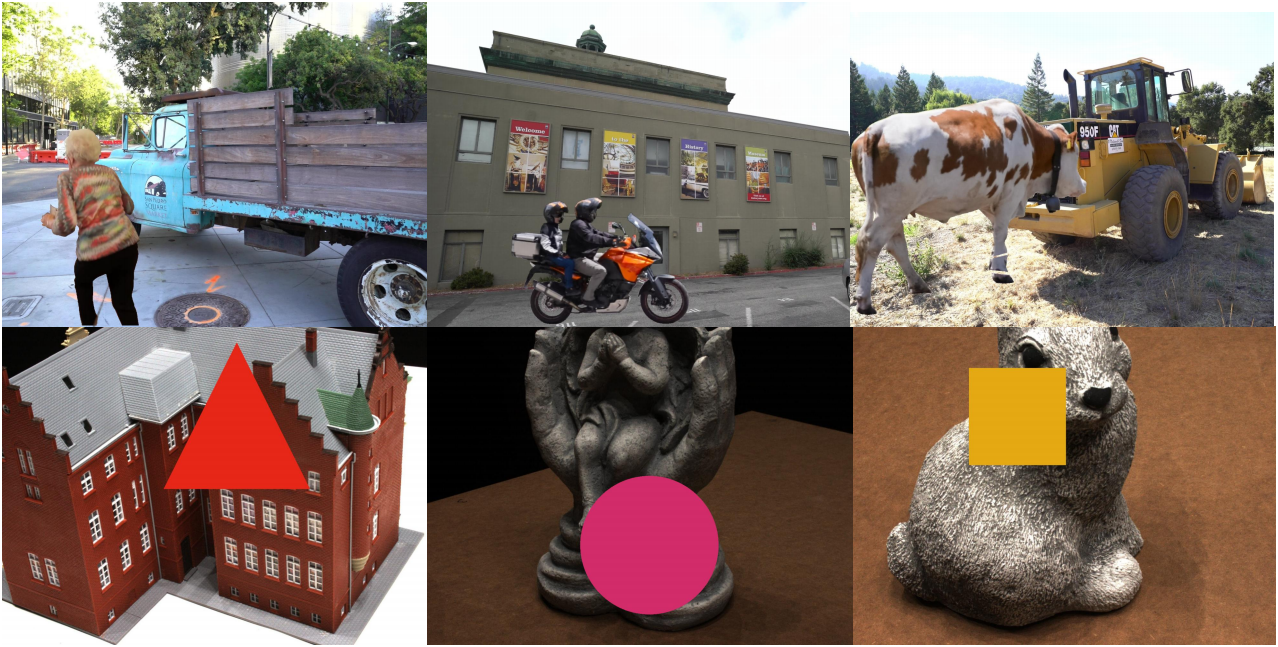}
    \caption{Example of distractors of DTU-Robust and TnT-Robust dataset.}
    \label{fig:dtu-robust_show}
\end{figure}

\begin{figure}
    \centering
    \includegraphics[width=\linewidth]{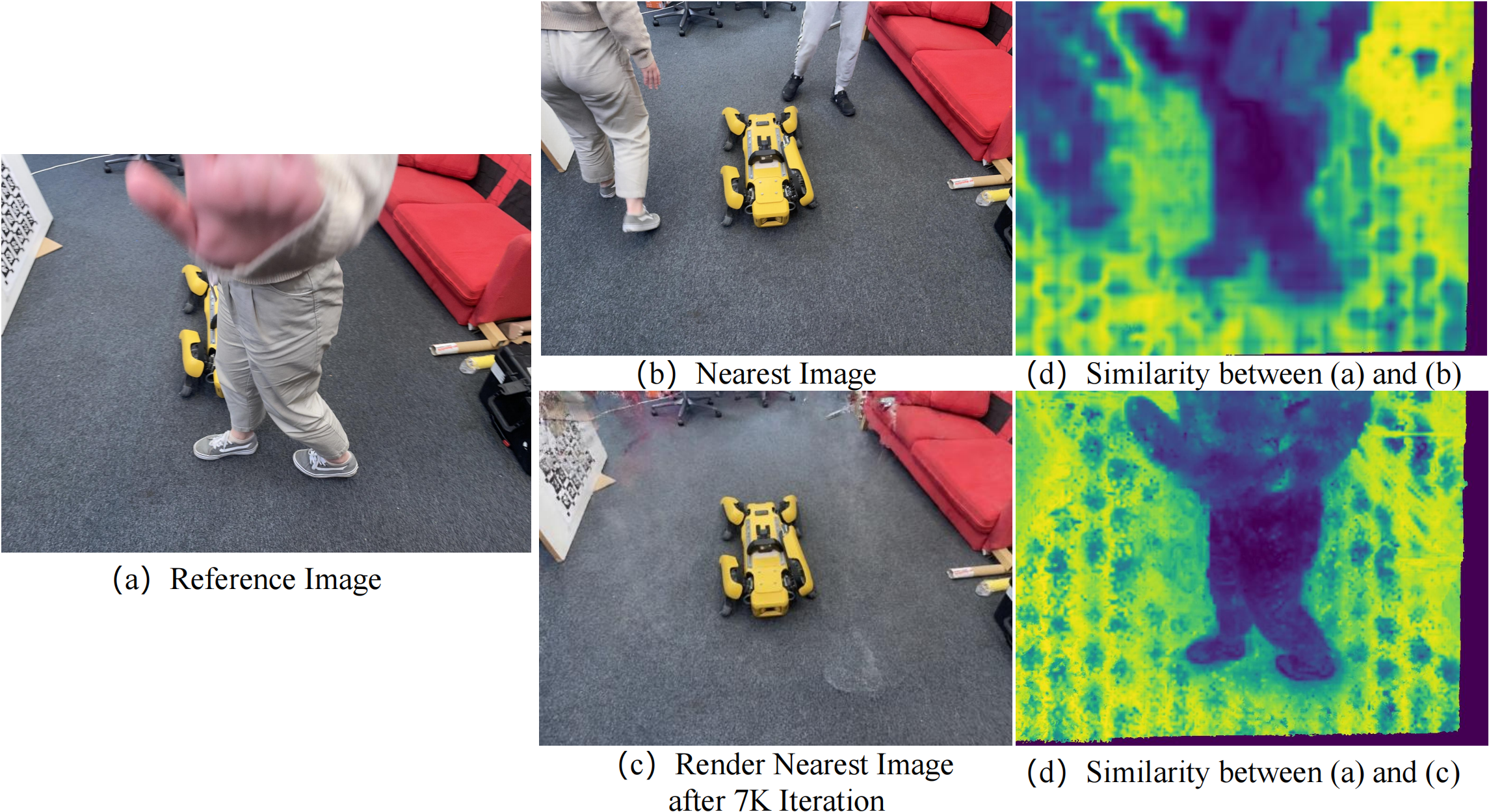}
    \vspace{-5mm}
    \caption{Reference image (a), nearest image after 7000 iteration (b), rendered image (c), feature similarity map between reference and nearest viewpoints (d), and feature similarity map between reference and rendered images (e). The rendered image can remove some distracting objects, preventing interference with the features of the reference image.}
    \label{fig:feat}
\end{figure}

\subsection{Distractor Detection}
\label{sec:method:distractor-detection}
Similar to the traditional 3D Gaussian Splatting methods~\cite{kerbl20233d,li2024geogaussian}, scenes containing distractors are processed in the same initialization stage. This stage leverages sparse point clouds and camera parameters generated from structure-from-motion (SfM) techniques. Using this initial setup, a rough scene reconstruction is performed, resulting in the computation of the initial depth map \( D \), surface normals \( N \), and an initial rendered image \( I^* \). This initialization provides a foundational representation of the scene, which is then refined in subsequent stages of the method to improve accuracy and handle outliers effectively.

Then, three steps are proposed to achieve an effective distractor detection strategy. First, the Gaussian surfels observed from the viewpoint \( {I}_r \) are associated with the corresponding surfels detected from neighboring viewpoints \( {I}_n \), based on the relative camera pose \( \mathbf{R}, \mathbf{t} \) between the two viewpoints. For a pixel \( \mathbf{p}_r \) on image \( I_r \), the corresponding pixel \( \mathbf{p}_n \) in a neighboring viewpoint \( I_n \) can be calculated using the homographic relationship in Eqn. \ref{eq:homography1}. 

\begin{equation}
\mathbf{H}_{nr} = \mathbf{K}_n \mathbf{R}_{nr} \left( \mathbf{I} + \frac{1}{d_r} \cdot \mathbf{t}_{rn} \mathbf{n}_r^\top \right) \mathbf{K}_r^{-1}
\label{eq:homography1}
\end{equation}
\begin{equation}
\mathbf{p}_{n} = \mathbf{H}_{nr} \mathbf{p}_{r}
\label{eq:homography2}
\end{equation}

This allows us to establish correspondences between pixels across different viewpoints, which is essential for identifying and handling potential distractors in the scene.

For a given reference image \( I_r \) and an initially rendered neighboring viewpoint \( I^* \), we use DINOv2\cite{oquab2023dinov2} to extract image features following the Nerf-on-the-Go\cite{ren2024nerf} method, resulting in feature maps \( F_r \) and \( F^*_n \). 
To remove the transient distracting objects that may also be present in the neighboring viewpoints which misleads the feature similarity calculation, we use the initially rendered image \( I^* \) instead of the original neighboring viewpoint image \( I_n \).
This issue is illustrated in Fig.~\ref{fig:feat}, where distracting objects in the original neighboring viewpoint image \( I_n \) interferes with accurate feature matching and distractor detection. 

Based on the two associated points, we estimate the feature distance between them shown in Eqn.~\ref{eq:dist}. The number of channels in the feature map is 384. To compute the feature vectors for two points on the corresponding feature maps, we first apply bilinear interpolation to obtain the features \( F_r(\mathbf{p}_r), F^*_n(\mathbf{p}_n) \), respectively. This is necessary because the feature map is downsampled by a factor of 14 from the original image, and bilinear interpolation allows us to extract the feature values at specific pixel locations. Pixels with a distance lower than a certain threshold \( \delta_{near} \) are used as a mask for the adjacent view. Then, the computation process of the distance is denoted as:
\begin{equation}
    distance(\mathbf{p}_r,\mathbf{p}_n) =  abs(\frac{F_r(\mathbf{p}_r) \cdot F^*_n(\mathbf{p}_n)}{\\|F_r(\mathbf{p}_r)\\|  
   \\|F^*_n(p_n)\\|}) 
    \label{eq:dist}
\end{equation}
where $ M_{n}$ is defined based on $distance_{rn}(\mathbf{p}_r) < \delta_{near}$. 

Second, each mask \( M_n \) obtained from the first step contains many incorrect segmentations due to rough geometric estimates and noise. Using the multi-view mapping relationships, we calculate the visibility of the 3D spatial points corresponding to the pixels classified as clutter in the mask and the current view. If a pixel is identified as clutter by at least two visible adjacent views, it is retained as the final segmentation result  \( Mask_{mv} \).

Third, the quality of segmentation in some boundary regions of the coarse mask is not very accurate.
Therefore, we continue to refine these regions based on the segment anything model (SAM)~\cite{kirillov2023segment} which is a prompt-based segmentation model. We perform uniform sampling on the $Mask_{mv}$ to obtain positive sample points on the clutter and negative sample points in the background region. These are computed with the original image using the outputs of SAM, resulting in a clutter mask \( Mask_{\text{sam}} \) with precise edges.
\begin{equation}
    Mask_{sam} = SAM(I_r, \mathcal{P,N} )
    \label{eq；m-sam}
\end{equation}

\subsection{Multi-view Pruning}
\label{sec:method:mulview-prune}
In the process of reconstructing clutter, the optimization often causes Gaussians to cluster near the camera. These floaters are typically incorrect models of viewpoint-dependent phenomena for clutter that only exists for a few frames. Once these floaters appear during optimization, they force the line of sight to reach the transmittance limit prematurely, preventing gradient propagation. 3DGS is to reset the opacity of all Gaussians every few iterations, using it as a control mechanism. This allows gradients to flow again and prunes Gaussians that cannot regain higher opacity.

However, in regions with clutter, resetting opacity is not fully effective. First, due to the presence of masks, this area tends to split into more Gaussians after opacity reset. For masks that cannot be perfectly segmented, this will lead to further aggravation of floaters. Therefore, we propose multi-view contribution-based pruning (MV-Prune). We define multi-view contribution as:
\begin{equation}
    \mathbf{C}_{MV}(p)=\sum_{\mathbb{V}_{k}\in {\mathbb{V}}} (\sum_{p \in \mathbb{V}_{k}} \alpha_{i(p)} \prod_{j=1}^{i(p)-1}\left(1-\alpha_{j}\right) > \delta_{opacity})
    \label{eq:mv-contri}
\end{equation}
where \( V_k \) is the training viewpoint, and \( p \) is the Gaussian for which contribution needs to be calculated. The contribution of Gaussian \( p \) for viewpoint \( V_k \) is measured by the cumulative transmittance of the pixel associated with Gaussian \( p \). When the cumulative transmittance along the ray exceeds the threshold \(\delta_{\text{opacity}}\), the contribution of Gaussian \( p \) in this viewpoint is incremented by one.

When the opacity of a Gaussian in a certain view exceeds a threshold, the contribution is raised by one. Once \( C_{\text{MV}}(p) \) exceeds the threshold $\delta_{prune} $, its transmittance is reset to a lower value to allow gradients to flow again. Experimental results show that MV-Prune effectively handles floaters, compressing by 60\% with comparable rendering quality.

\subsection{Multi-view Consistency}
\label{sec:method:mulview-loss}
To enhance geometric consistency, we use photometric consistency constraints based on neighboring patches, similar to MVS algorithms. We use a homography matrix to compute the $11\times 11$ pixel patch \(P_r\) around pixel \(\mathbf{p}_r\), mapping it to the corresponding region \(P_n\) in the neighboring view. To measure consistency, we use the normalized cross correlation (NCC)~\cite{Yoo2009FastNC} coefficient as a loss metric:
\begin{equation}
    \boldsymbol{L}_{mv}=\frac{1}{V} \sum_{\boldsymbol{p}_{r} \in V}\left(1-N C C\left(\boldsymbol{I}_{r}\left(\boldsymbol{p}_{r}\right), \boldsymbol{I}_{n}\left( \boldsymbol{p}_{n}\right)\right)\right).
    \label{eq:mv-loss}
\end{equation}

To focus on these areas with reconstruction errors, we calculate per-pixel geometric reconstruction accuracy weights. This is done by reprojecting the corresponding pixel \(\textbf{p}_n\) from the neighboring view back using the homography matrix to obtain \(\mathbf{p}^{'}_n\), then calculating the reprojection error and weight. This weight can also handle out-of-bounds and potential occlusion issues in different views. 
\begin{equation}
 \boldsymbol{E}_{repro}=\frac{1}{V} \sum_{\boldsymbol{p}_{r} \in V} 
 \left\|\boldsymbol{p}_{r}-\boldsymbol{H}_{rn }\boldsymbol{p}_{n}\right\| 
    \label{eq:mv-loss2}
\end{equation}
\begin{equation}
\boldsymbol{w}_{repro}=\frac{1}{1+\boldsymbol{E}_{repro}}
    \label{eq:mv-loss3}
\end{equation}

Unlike MVS algorithms, we do not directly optimize the reprojection error, as the lack of neighborhood information can impair the gradient propagation of Gaussian errors. Using color error alone is sufficient for depth optimization.

Another loss function is the regularization term \(L_s\) that minimizes the shortest axis, modeling the Gaussian as a thin surface. The image reconstruction loss is $L_{rgb}$. In summary, the loss function is:
\begin{equation}
\boldsymbol{L}=L_{rgb}+\lambda _{1}L_{s}+\lambda _{2}w_{repro}L_{mv}
    \label{eq:loss}
\end{equation}

 For the surface loss ,We set $\lambda _{1}$ = 100. For the geometric loss, we set $\lambda _{2}$ = 0.2.
\label{sec:method}

\section{Experiments}
\label{sec:experiments}
In this section, we present both qualitative and quantitative comparisons of our method with state-of-the-art Gaussian Splatting reconstruction approaches~\cite{huang20242d, chen2024pgsr} on public datasets.

\subsection{Implementation}
All experiments were conducted on a single NVIDIA 4090 GPU. The maximum number of nearest images in multi-view scenarios was set to $8$, with a maximum angular difference of $60$ degrees between adjacent view cameras and a maximum distance of $1.5$. The novel view rendering task was performed using the training results from $7,000$ iterations, and the corresponding relationships between adjacent views were calculated using Eqn.~\ref{eq:dist}. When computing the multi-view masks in Sec. \ref{sec:method:distractor-detection}, the cosine similarity threshold $\delta_{near}$ is set to $0.5$. For SAM prompt points, $20$ segmentation prompt points and $1$ exclusion prompt point are used, with the segmentation results being considered valid if the top $10$ votes exceeded $2$, serving as the refined mask segmentation results. The masks are then used for re-training, resulting in high-quality surface reconstruction after $30,000$ iterations. In the multi-view pruning phase, the transmittance contribution was set to $0.5$, and only views with contributions exceeding the threshold of $8$ were retained.

\begin{figure*}[htbp]
	\centering
    \includegraphics[width=0.9\linewidth]{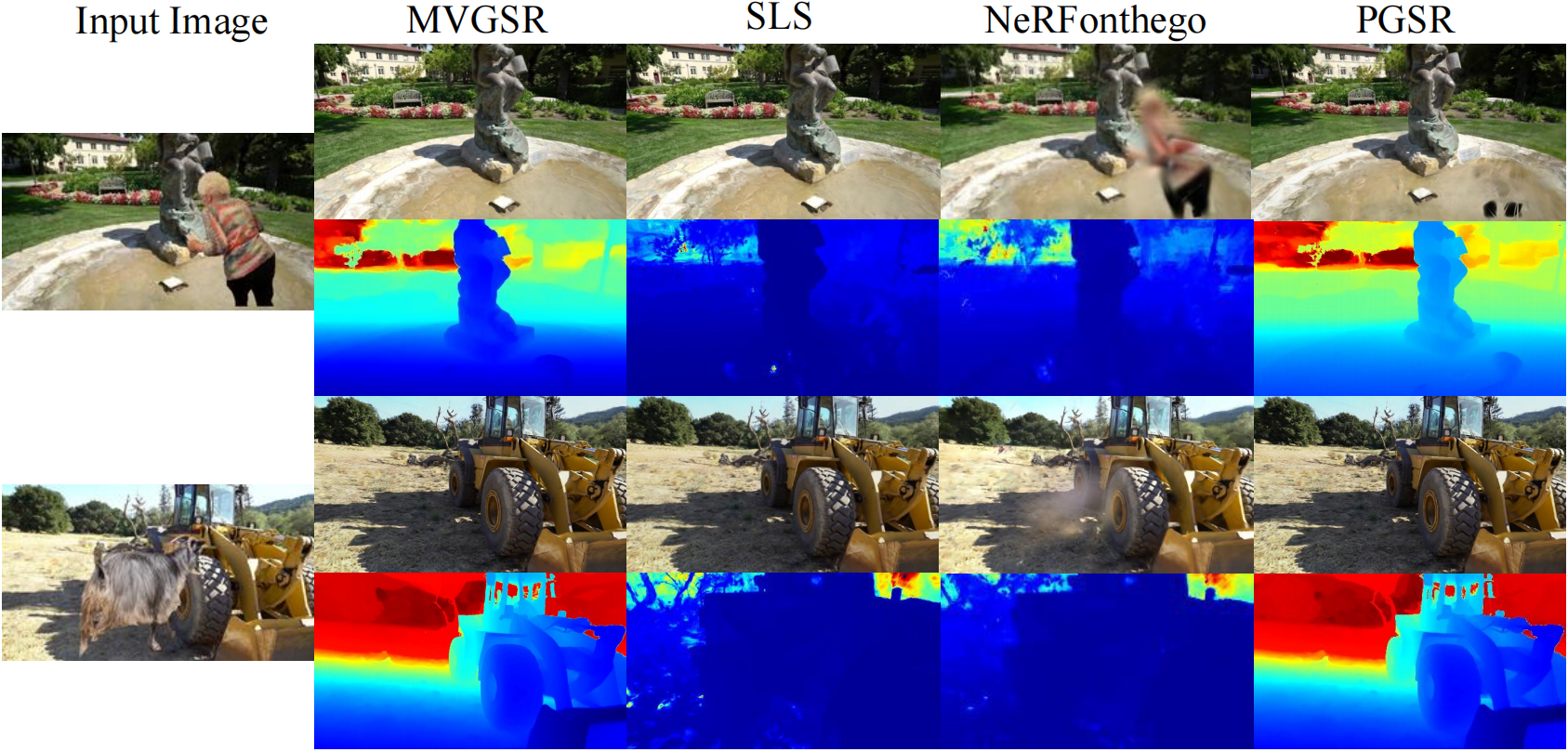}
\caption{Comparison of view rendering on the TnT-Robust dataset.  }
\label{fig:compare_tnt-robust}
\end{figure*}

\begin{table*}
\centering
\resizebox{0.8\textwidth}{!}{
\begin{tabular}{cc|c|c|c|c|c|c|c}
\toprule
\multicolumn{2}{c|}{scan} & Truck & Caterpillar & Barn & Meetingroom & Ignatius & Courthouse & Avg. \\ \cline{1-9}
\multirow{4}{*}{F1$\uparrow$} 
&PGSR~\cite{chen2024pgsr} 
& 0.57 & 0.37 & {0.57} & 0.30 & 0.64 & {0.18} & 0.44 \\ \cline{2-9}
&SLS~\cite{sabourgoli2024spotlesssplats} 
& 0.48 & 0.30 & 0.43 & 0.24 & 0.57 & 0.15 & 0.36 \\ \cline{2-9}
&NeRFonthego~\cite{Ren2024NeRFonthego} 
& 0.37 & 0.22 & 0.28 & 0.17 & 0.49 & 0.11 & 0.27  \\ \cline{2-9}
&MVGSR 
& {0.60} & {0.42} & {0.59} & {0.34} & {0.73} & 0.18 & {0.48} \\

\bottomrule
\end{tabular}
}
\caption{Quantitative results of reconstruction performance on the TnT-Robust dataset.}
\label{table_tnt}
\end{table*}

\begin{table*}
\centering
\resizebox{\textwidth}{!}{
\begin{tabular}{cc|c|c|c|c|c|c|c|c|c|c|c|c|c|c|c|c}
\toprule
\multicolumn{2}{c|}{scan} & 24 & 37 & 40 & 55 & 63 & 65 & 69 & 83 & 97 & 105 & 106 & 110 & 114 & 118 & 122 & Avg. \\ \cline{1-18}
\multirow{3}{*}{CD$\downarrow$} 
&PGSR~\cite{chen2024pgsr} 
& 0.53 & 1.00 & 0.51 & 0.36 
& 1.14 & 0.61 & 0.49 & 0.91 
& 0.62 & 0.59 & 0.47 & 0.46 
& 0.32 & 0.37 & 0.35 & 0.61 \\ \cline{2-18}
&2DGS~\cite{huang20242d} 
& 0.52 & 0.86 & 0.60 & 0.45 
& 1.12 & 0.99 & 0.82 & 1.37 
& 1.19 & 0.69 & 0.71 & 0.70 
& 0.41 & 0.70 & 0.56 & 0.77 \\ \cline{2-18}
&MVGSR 
& \textbf{0.34} & \textbf{0.51} & \textbf{0.30} & \textbf{0.30} 
& \textbf{0.44} & \textbf{0.52} & \textbf{0.45} & \textbf{0.63} 
& \textbf{0.59} & \textbf{0.56} & \textbf{0.35} & \textbf{0.38} 
& \textbf{0.29} & \textbf{0.34} & \textbf{0.35} & \textbf{0.42} \\ \cline{1-18}
\multirow{3}{*}{PSNR$\uparrow$} 
&PGSR~\cite{chen2024pgsr} 
& 31.34 & 26.72 & 29.88 & 31.81 
& 32.66 & 31.27 & 30.94 & 32.02 
& 30.37 & 32.76 & 34.94 & 33.58 
& 32.16 & 36.41 & 35.71 & 32.15 \\ \cline{2-18}
&2DGS~\cite{huang20242d} 
& 30.31 & 28.73 & 27.87 & 26.65 
& 33.52 & 33.66 & 31.09 & 31.23 
& 33.22 & 30.99 & 30.89 & 32.31 
& 32.16 & 33.58 & 35.87 & 31.47 \\ \cline{2-18}
&MVGSR 
& \textbf{33.06} & \textbf{29.22} & \textbf{32.52} & \textbf{34.29} 
& \textbf{37.01} & \textbf{34.11} & \textbf{32.86} & \textbf{40.90} 
& \textbf{34.37} & \textbf{38.02} & \textbf{37.96} & \textbf{35.81} 
& \textbf{33.21} & \textbf{39.40} & \textbf{39.82} & \textbf{35.58} \\ \bottomrule
\end{tabular}
}
\caption{Quantitative results of reconstruction performance on the DTU-Robust dataset. \textbf{Bold} indicates the best results.}
\label{table_2_recon_CD}
\end{table*}

\begin{figure*}[htbp]
	\centering

	\begin{minipage}{0.19\linewidth}
		\centering
        \text{Ground Truth} \\[0.1cm]
       
		\includegraphics[width=\linewidth]{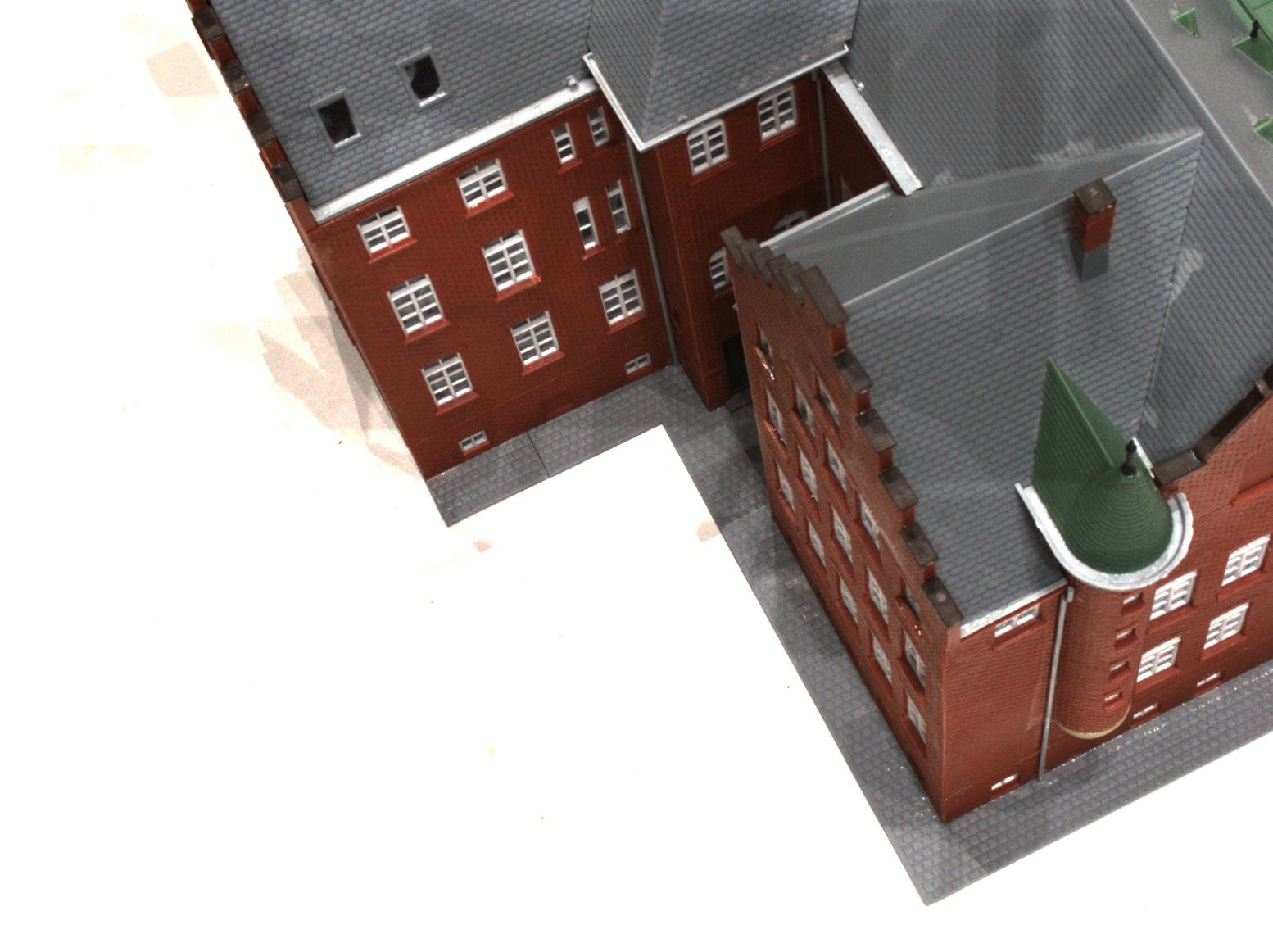}
	\end{minipage}
	\begin{minipage}{0.19\linewidth}
		\centering
        \text{Ours} \\[0.1cm]
		\includegraphics[width=\linewidth]{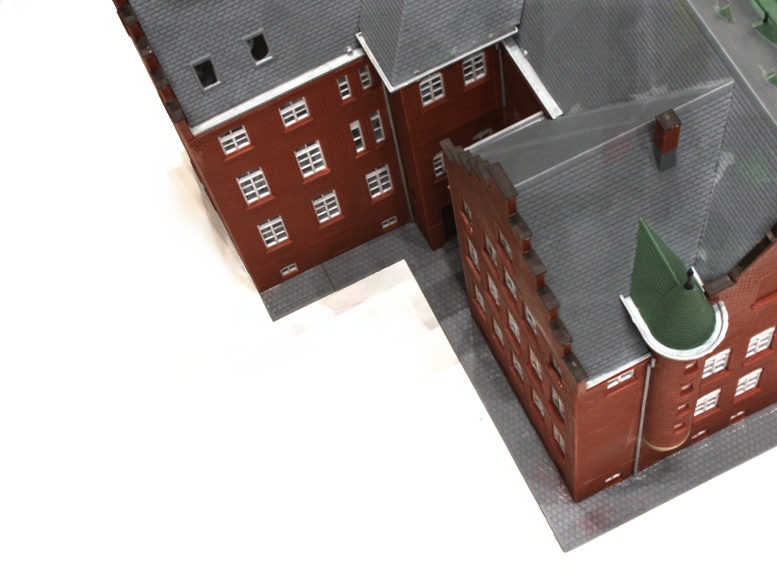}
	\end{minipage}
	\begin{minipage}{0.19\linewidth}
		\centering
        \text{3DGS} \\[0.1cm]
		\includegraphics[width=\linewidth]{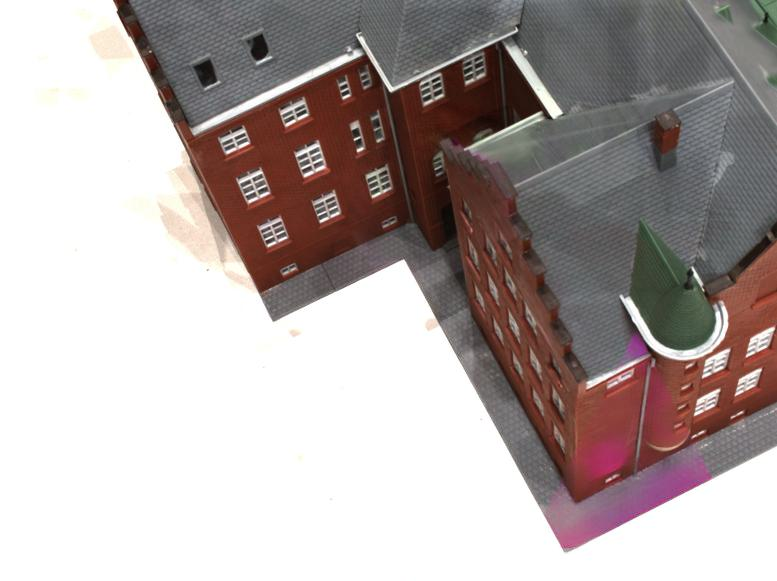}
	\end{minipage}
    \begin{minipage}{0.19\linewidth}
		\centering
        \text{2DGS} \\[0.1cm] 
		\includegraphics[width=\linewidth]{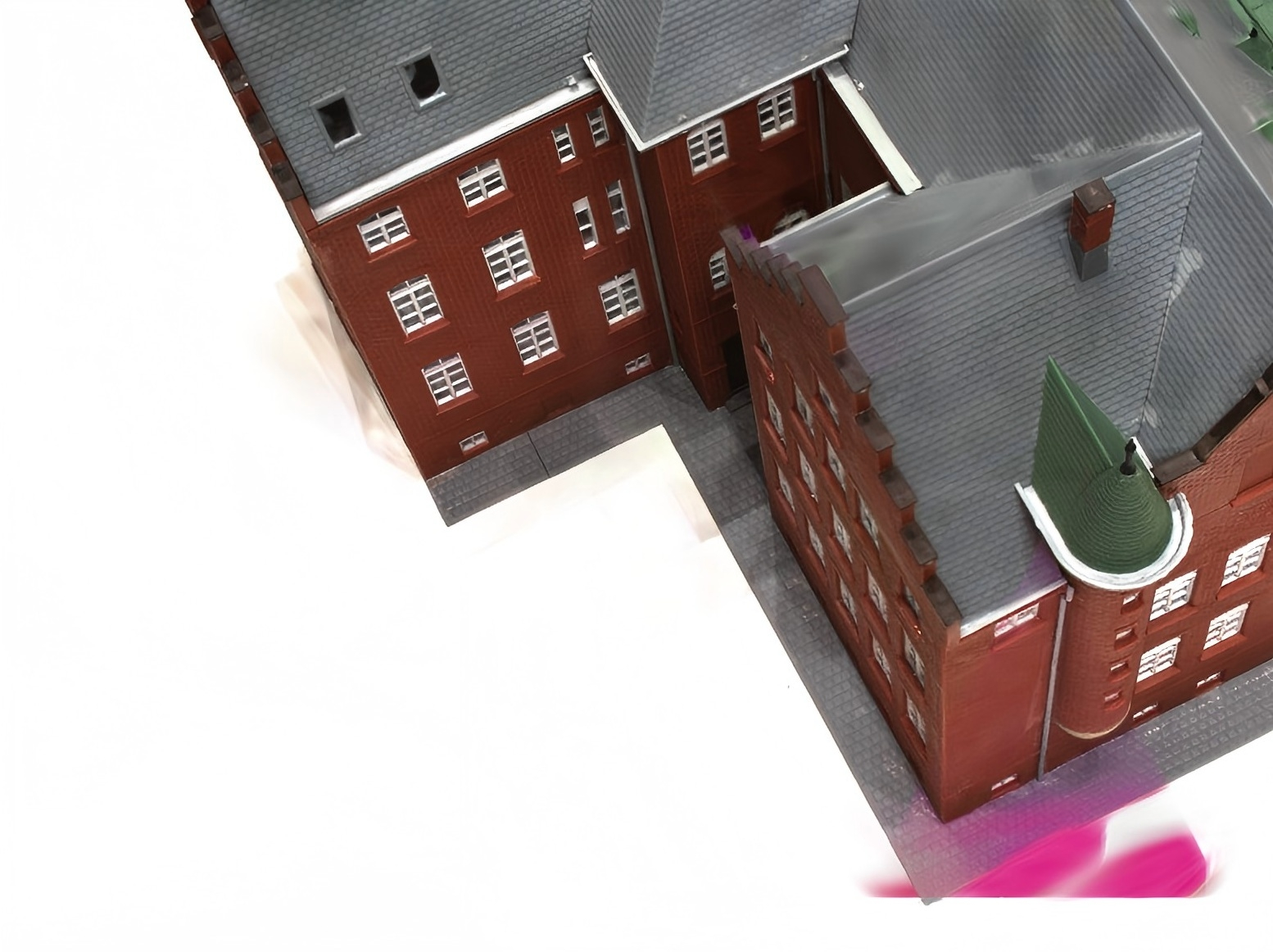}
	\end{minipage}
	\begin{minipage}{0.19\linewidth}
		\centering
        \text{PGSR} \\[0.1cm] 
		\includegraphics[width=\linewidth]{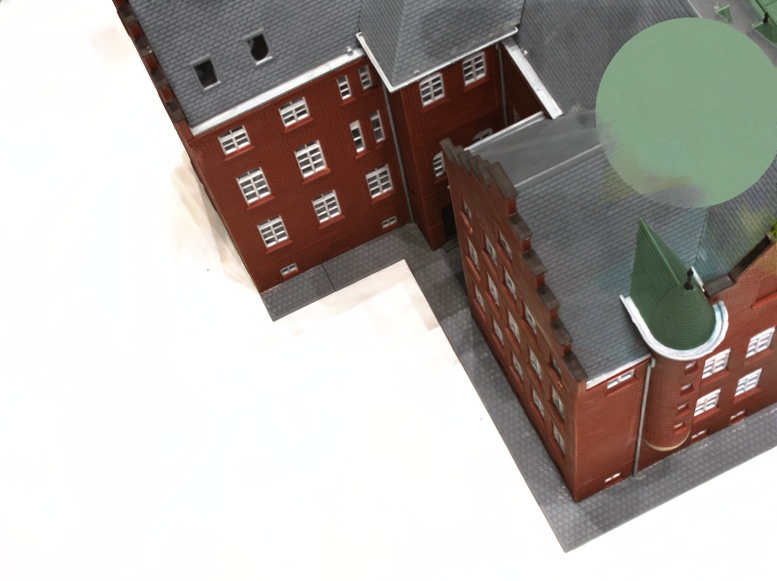}
	\end{minipage}
	
	\begin{minipage}{0.19\linewidth}
		\centering
		\includegraphics[width=\linewidth]{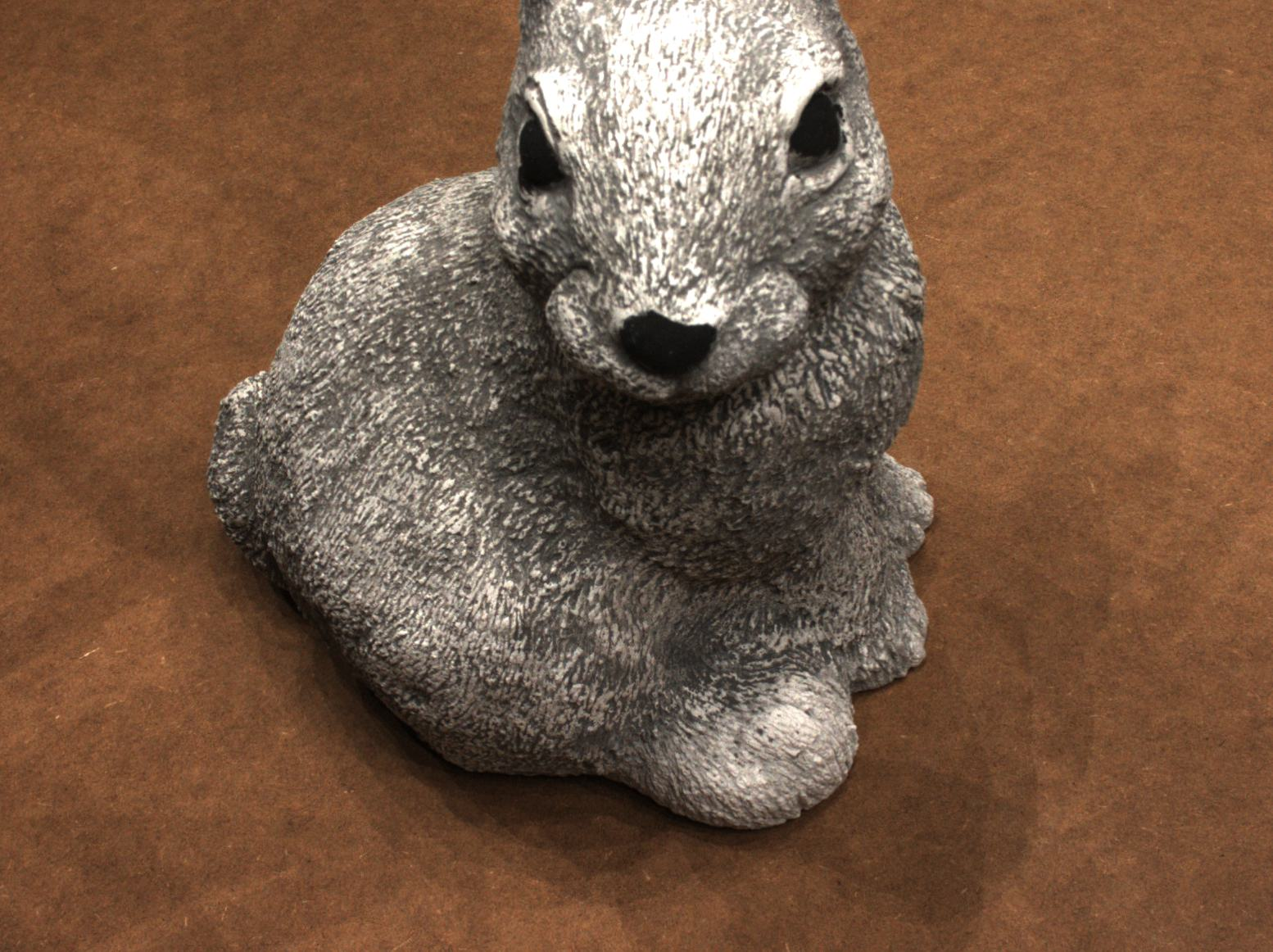}
	\end{minipage}
	\begin{minipage}{0.19\linewidth}
		\centering
		\includegraphics[width=\linewidth]{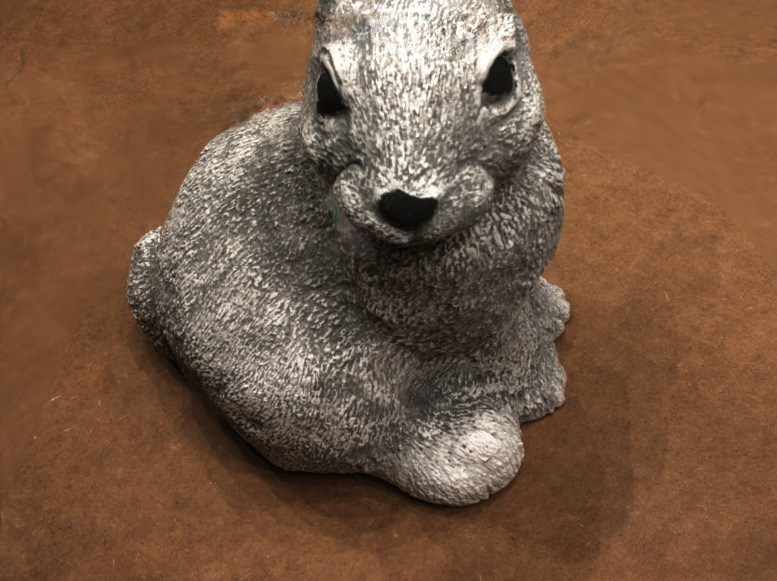}
	\end{minipage}
	\begin{minipage}{0.19\linewidth}
		\centering
		\includegraphics[width=\linewidth]{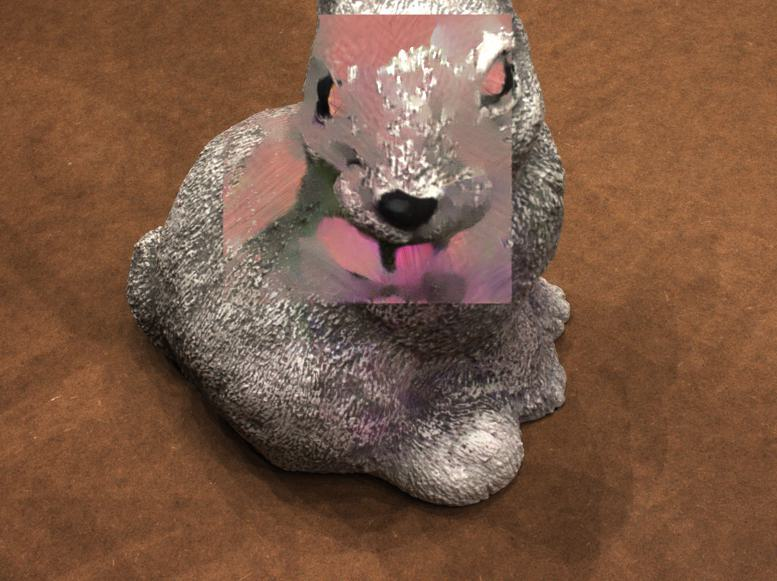}
	\end{minipage}
    \begin{minipage}{0.19\linewidth}
		\centering
		\includegraphics[width=\linewidth]{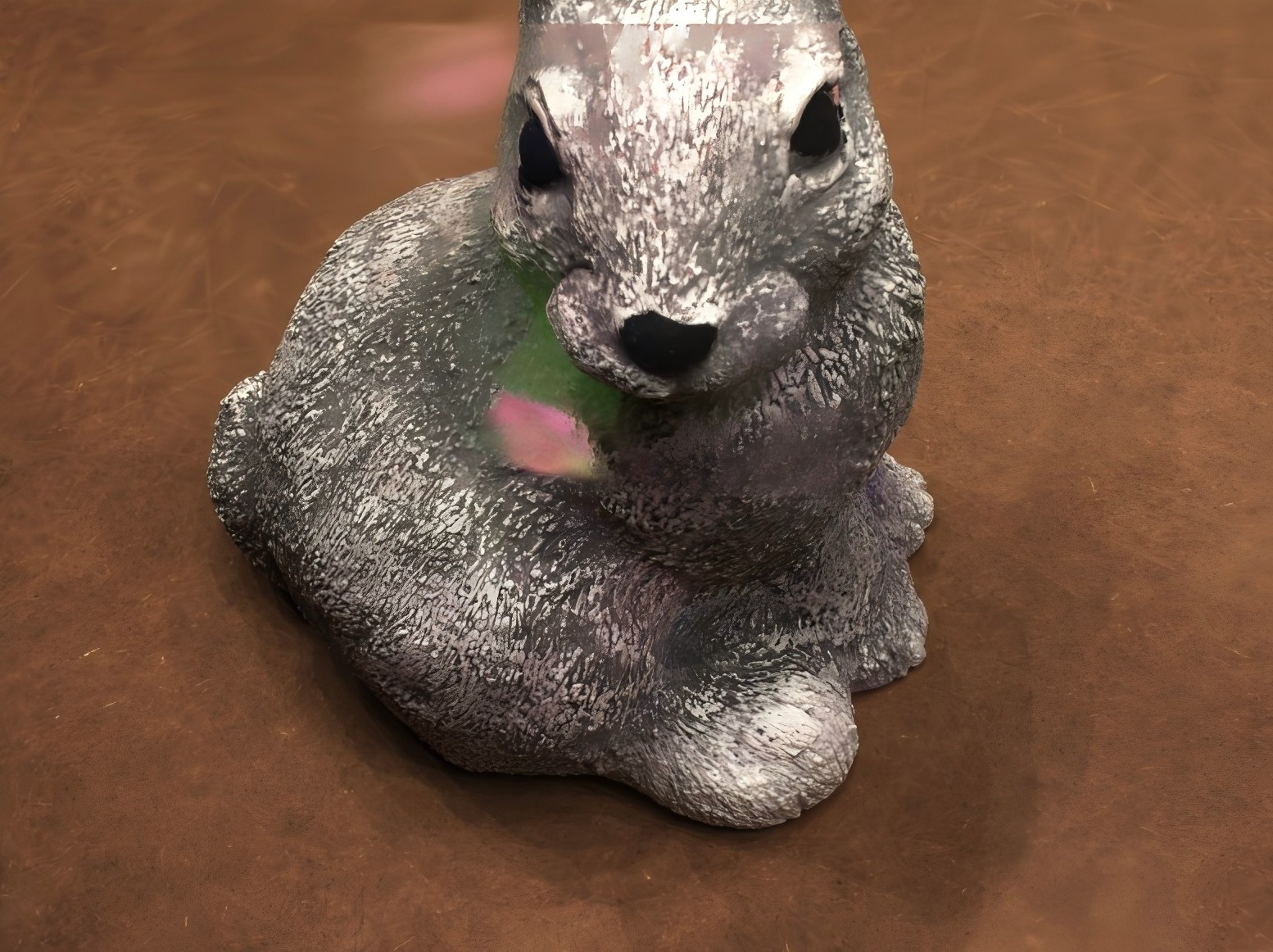}
	\end{minipage}
	\begin{minipage}{0.19\linewidth}
		\centering
		\includegraphics[width=\linewidth]{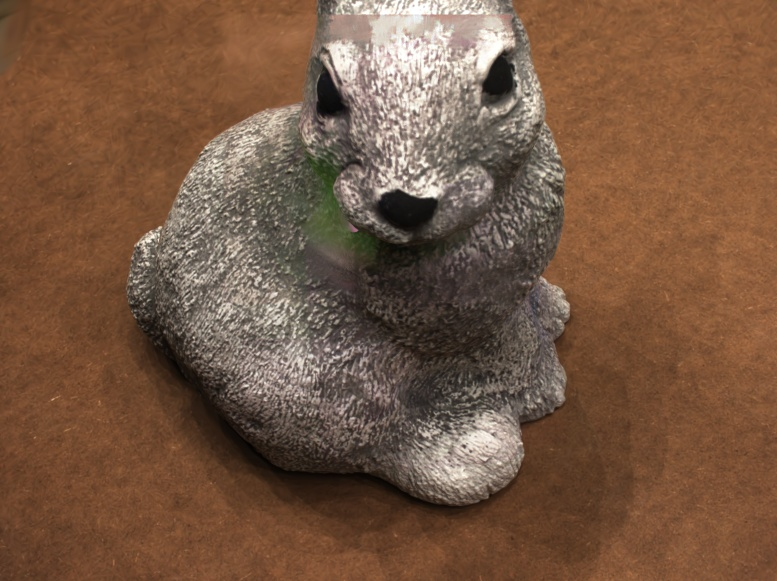}
	\end{minipage}

\caption{Comparison of view rendering on the DTU-Robust dataset. We select scan scene  24 (building), 55 (rabbit)}
\label{fig:compare_dtu-robust}
\end{figure*}
\begin{figure}
	\centering
	\includegraphics[width=\linewidth]{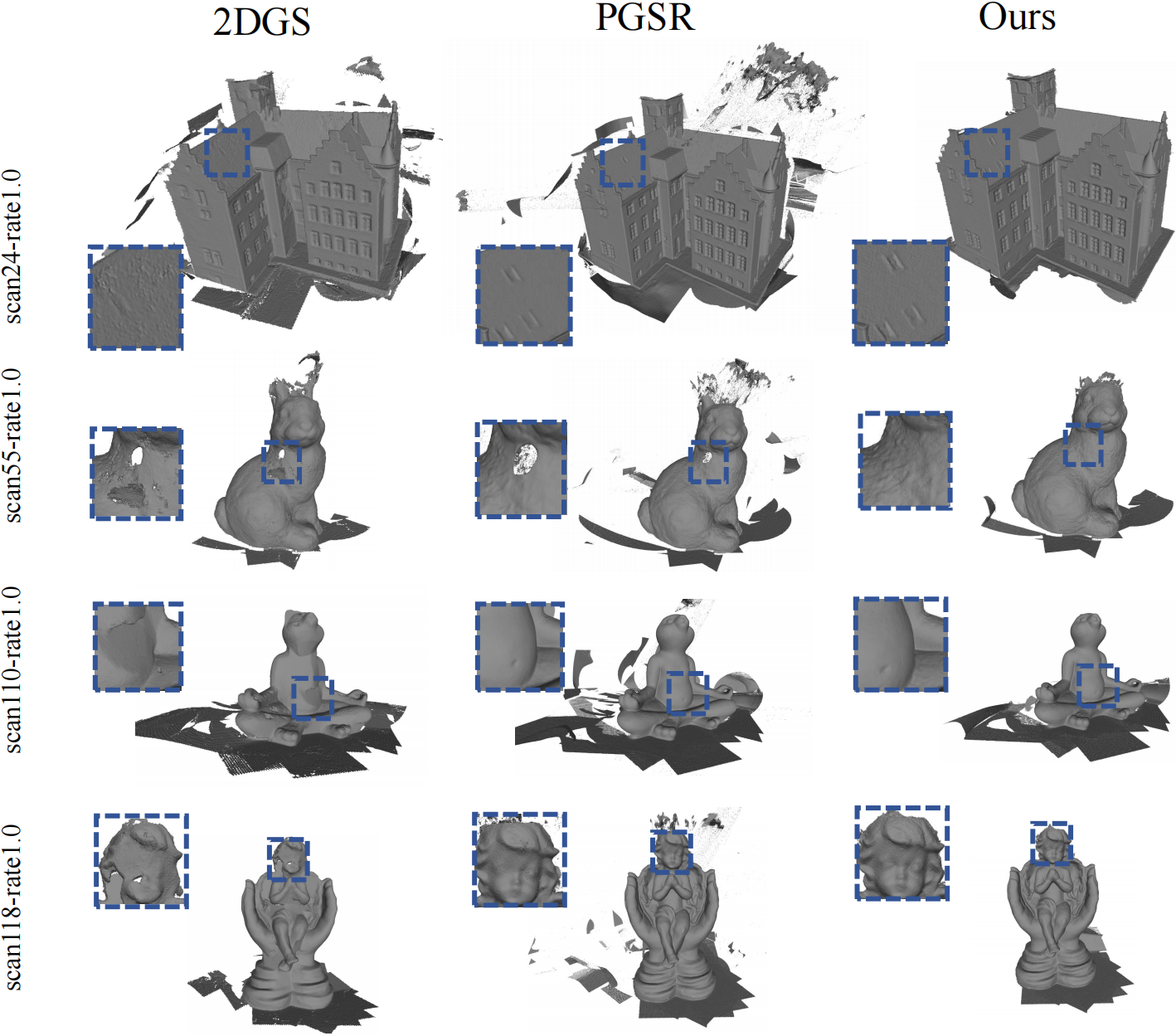}
\vspace{-6mm}
\caption{Comparison of reconstruction on DTU-Robust dataset.}
\vspace{-4mm}
\label{fig:compare_dtu-3d-robust}
\end{figure}

\subsection{Datasets and Metrics}
 \paragraph{Datasets.}To evaluate reconstruction and rendering performance, we employ the DTU~\cite{jensen2014large} and TnT~\cite{tnt} datasets with ground truth 3D models, containing 15 object-centric scenes and 6 large-scale outdoor scenarios respectively. To address their limitation in dynamic elements, we introduce DTU-Robust and TnT-Robust (publicly released), incorporating distractors as shown in Fig.~\ref{fig:dtu-robust_show}.

In DTU-Robust, random geometric shapes (squares, circles, triangles) of varying sizes are superimposed on training images at rates $r = {0.3, 0.5, 0.8, 1.0}$ to control distractor density. TnT-Robust enhances real-world fidelity by injecting distractors (pedestrians, vehicles, animals) from DAVIS 2017~\cite{davis2017} into static scenes. Both extended datasets enable comprehensive robustness testing under distractors while preserving original reconstruction benchmarks.

 \paragraph{Metrics for rendering and reconstruction evaluations.} 
 In our experiments, rendering quality is measured via Peak Signal-to-Noise Ratio for novel views. Surface reconstruction uses Chamfer Distance (for DTU) and F1 score (for TnT), as is commonly done following the methods in this area. The bidirectional CD evaluates point cloud alignment, while F1 quantifies geometric consistency by balancing precision and recall through threshold-based surface distance analysis.

\subsection{Reconstruction}
 In this section of the paper, we evaluated the reconstruction capabilities of the distractor-free algorithms SLS\cite{sabour2024spotlesssplats} and NeRF-Onthego\cite{Ren2024NeRFonthego}, as well as the surface reconstruction algorithm PGSR\cite{chen2024pgsr}, on the TnT-Robust dataset. As shown in Table \ref{table_tnt}, on TnT-Robust, we outperformed PGSR in F1 scores by 0.03 (Truck), 0.05 (Caterpillar), and 0.09 (Ignatius), highlighting robustness against dynamic interference. On DTU-Robust, our method achieved a mean CD score of 0.42, surpassing 2DGS and PGSR by 76.45\% and 60.31\% on average, with consistent performance across occlusion rates and scenes.  

 Next, we present a qualitative comparison with the baselines on the DTU-Robust dataset Our method demonstrates superior reconstruction quality across both DTU-Robust and TnT-Robust datasets compared to baseline models (2DGS, PGSR, SLS, NeRFonthego). On DTU-Robust (Fig.~\ref{fig:compare_dtu-3d-robust}), our outputs feature smoother surfaces with fewer holes/distractors, notably resolving complex geometries like the rabbit neck (scan 55) without fragmentation. In contrast, 2DGS loses fine details, while PGSR introduces scattered artifacts. For TnT-Robust (Fig.~\ref{fig:compare_tnt-robust}), our multi-view consistent representation ensures geometric continuity, outperforming SLS (distractor-free but artifact-prone) and PGSR (discontinuous surfaces). Key advantages include detail preservation, artifact suppression, and robustness to challenging scenes. This highlights our method’s ability to balance structural accuracy with surface coherence under diverse real-world conditions.

\subsection{\textcolor{black}{View rendering}}
We demonstrate the rendering performance on the DTU-Robust dataset and compare it with PGSR and 2DGS. As shown in Table \ref{table_2_recon_CD}, our method consistently achieves higher PSNR across different scenes. Please note that the values here are the average results at different rates, verifying the stability of the performance in random scenes. Particularly in the scan83 scene, our method outperforms 2DGS and PGSR by 9.67 dB and 8.88 dB in PSNR, respectively. On average, our method achieves 35.58 dB in PSNR and demonstrates superior performance compared to the baseline methods. The higher results across scenes indicate that our method exhibits robustness and strong performance in novel view synthesis. 

We evaluated our method qualitatively in comparison to other competing methods as shown in Fig.~\ref{fig:compare_dtu-robust}. Our method clearly removes most distractors and distortion compared to the other methods. Our method clearly resembles the ground truth the closest without distractors and this validates the effectiveness of our pruning strategy in alias removing. For results on additional datasets, please refer to the supplementary materials.

\subsection{Ablation Study}

To demonstrate the contribution of each component, ablation experiments were conducted on DTU-Robust scan24 in Table \ref{table:ablation2}. The multi-view Loss $L_{mv}$ significantly improves surface reconstruction quality by reducing the Chamfer Distance  from 1.63 to 0.37, despite a slight decrease in rendering performance. Mask and Mv-Prune effectively reduce artifacts, ultimately achieving high-fidelity rendering with superior quality preservation. Also, we compare the runtime and resource consumption of different algorithms on scan24 of DTU-Robust dataset. For more details, please refer to the supplementary materials.

\begin{table}[h]
\vspace{-2mm}
\centering
\resizebox{\linewidth}{!}{
\begin{tabular}{cccc|cc}
\cline{1-6}
mask\_mv    & mask\_sam  & L\_mv     & MV-Prnue & PSNR↑                & CD↓                  \\ \hline
\checkmark  & \checkmark &             & & 32.20                & 1.63                 \\
\checkmark  &            & \checkmark  & & 30.01                & 0.56                 \\
\checkmark  & \checkmark & \checkmark  & & 30.75                & 0.37 \\ 
\checkmark  & \checkmark & \checkmark  &  \checkmark  & 33.06                & 0.34 \\ \hline               
\end{tabular}
}


\caption{Ablation Study}
\vspace{-5mm}
\label{table:ablation2}
\end{table}


\section{Discussion and Conclusion}
\label{sec:conclusion}
In this paper, we introduced Multi-View Consistency Gaussian Splatting for Robust Surface Reconstruction (MVGSR), a novel approach that addresses the common issues of floating artifacts and color errors in 3D Gaussian Splatting when applied to surface reconstruction. While 3DGS has gained popularity due to its high-quality rendering and fast training speeds, it is prone to distortions from distractors across different viewpoints, which can severely affect the accuracy and quality of surface reconstruction. Our method mitigates these issues by focusing on multi-view consistency to identify and separate distractors from the static elements in the scene.

{
    \small
    \bibliographystyle{ieeenat_fullname}
    \bibliography{main}
}
\appendix
\newpage
\setcounter{figure}{8}
\setcounter{table}{3}

In the supplementary materials, we analyze the effects of different masking strategies (section \ref{mask}), visualize the performance of MV-Pruned (section \ref{mvprune}), and report its quantitative metrics. We additionally provide ablation studies on key parameters and compare the computational overhead across algorithms (section \ref{ablation}). Extended experimental results on the DTU-Robust (section \ref{dtu}), TnT-Robust (section \ref{tntrobust}) and NeRF-onthego (section \ref{nerfonthego})datasets are also included for comprehensive evaluation.

\begin{figure}[h]
    \centering
    \includegraphics[width=\linewidth]{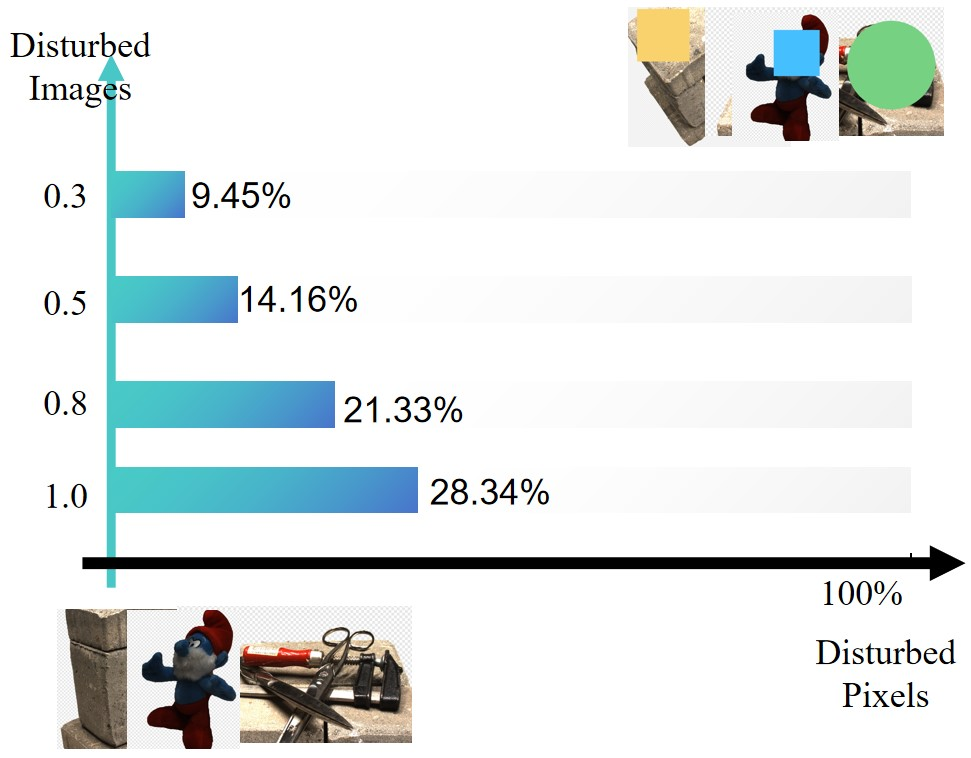}
    \caption{The average ratio of disturbed pixels to all object pixels under different disturbed image proportions on the DTU-Robust dataset.}
    \label{fig:dtu-robust}
\end{figure}
\begin{table}[h]
\centering
\resizebox{\linewidth}{!}{
\begin{tabular}{c|rccc}
\toprule
\multicolumn{1}{c|}{Dataset}  & Metric & 2DGS~\cite{huang20242d}  & PGSR~\cite{chen2024pgsr}  & Ours  \\ \hline
\multirow{3}{*}{fountain} & PSNR$\uparrow$   & 20.74 & 20.69 & 21.82 \\
                          & SSIM$\uparrow$   & 0.583 & 0.657 & 0.659 \\
                          & LPIPS$\downarrow$  & 0.424 & 0.353 & 0.355 \\ \hline
\multirow{3}{*}{corner}   & PSNR$\uparrow$   & 26.36 & 25.17 & 25.84 \\
                          & SSIM$\uparrow$   & 0.778 & 0.799 & 0.799 \\
                          & LPIPS$\downarrow$  &0.369  & 0.325 & 0.338 \\ \hline
\multirow{3}{*}{spot}     & PSNR$\uparrow$   & 22.57 & 22.72 & 22.99 \\
                          & SSIM$\uparrow$   & 0.576 & 0.593 & 0.586 \\
                          & LPIPS$\downarrow$  & 0.415 & 0.369 & 0.379 \\ \hline
\multirow{3}{*}{Avg.}     & PSNR$\uparrow$   & 23.22 & 22.86 & 23.55 \\
                          & SSIM$\uparrow$   & 0.645 & 0.683 & 0.681 \\
                          & LPIPS$\downarrow$  & 0.402 & 0.349 & 0.357 \\   
                          \bottomrule
\end{tabular}}
\caption{Comparison of novel view synthesis for \textcolor{black}{Gaussian Splatting reconstruction methods} on NeRF-on-the-go~\cite{ren2024nerf}.}
\label{on-the-go-recon}
\end{table}

\section{Pruning Analysis} 
\label{mvprune}
Quantitatively, we compare the effectiveness of our MV-Prune in Tab.~\ref{tab: mv-prune}. With MV-Prune, the storage requirement of the model is decreased by 30.19\% on average while maintaining 99.4\% of the original performance according to PSNR$\uparrow$ and a 0.01 decrease in CD$\downarrow$.

\begin{figure}
    \centering
    \begin{subfigure}{0.49\linewidth}
        \centering
        \includegraphics[width=\linewidth]{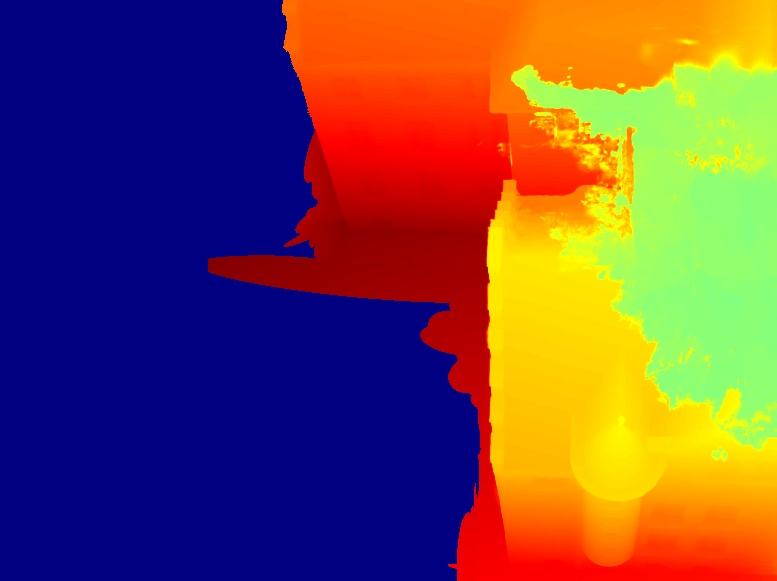}
        \caption{Before MV-prune }
        \label{fig:alias}
    \end{subfigure}
    \hfill
    \begin{subfigure}{0.49\linewidth}
        \centering
        \includegraphics[width=\linewidth]{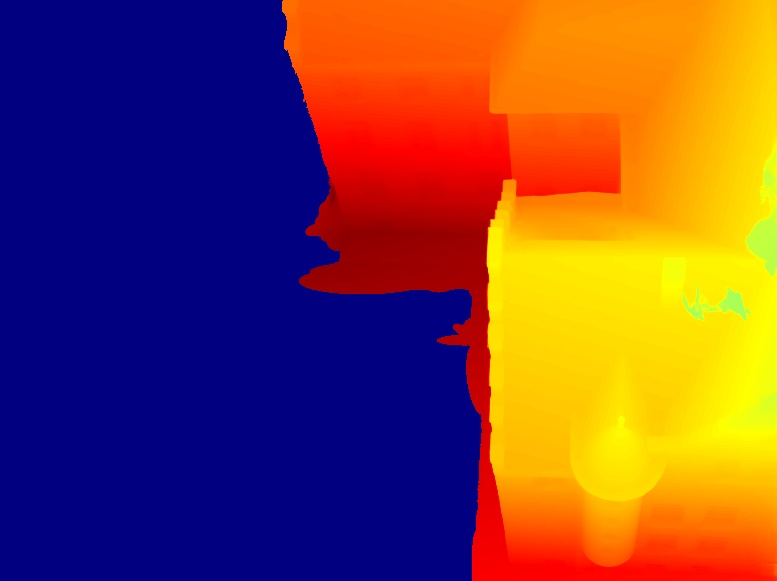 } 
        \caption{After MV-prune}
        \label{fig:second-image}
    \end{subfigure}
    \caption{Proposed pruning strategy MV-prune improves artifact removal, comparing results before and after MV-prune application.
    }
    \label{fig:mv-prune_comparison}
\end{figure}

Qualitatively, when we compare the images visualized in Fig.~\ref{fig:mv-prune_comparison}, we observe that the image on the right is cleaner with fewer distractors. Firstly, the green artifact present on the right of Fig.~\ref{fig:alias} is largely removed in Fig.~\ref{fig:second-image} with only a few small specks remaining. Moreover, the dark red jagged edges in the middle of Fig. \ref{fig:alias} are smoothed to become less pronounced in Fig.~\ref{fig:second-image}. Lastly, the details in the scene such as the structure of the object in yellow to orange are preserved in both pictures. These three observations show the efficacy of pruning with fewer distractors and better outlier removal whilst maintaining the high-frequency details of the scene.

\begin{table}[]
\centering
\resizebox{\linewidth}{!}{
\begin{tabular}{c|ccc|ccc}
\toprule
& \multicolumn{3}{c|}{MVPSR }                            & \multicolumn{3}{c}{MVGSR(+w MV-Prune)}                    \\ \cline{2-7}
& PSNR$\uparrow$& CD$\downarrow$& SIZE(MB)&PSNR$\uparrow$ & CD$\downarrow$ & SIZE(MB)
\\ \hline

0.0  & 35.71& 0.37 & { 27.28} & 38.49 & 0.35 & {28.00} \\ \cline{1-1}
0.3 & 35.85& 0.37 & 55.78 & {38.29} & 0.35& 33.02    \\ \cline{1-1}
0.5 & 35.95& 0.36& 56.74 & {38.59}  & {0.35} &35.45  \\ \cline{1-1}
{0.8}& 36.16& 0.39& 60.78 & {37.99} & {0.35} &35.57   \\ \cline{1-1}
{1.0}& 35.98 & 0.39 & 105.63 & 36.98 & 0.37& {34.90}   \\ \cline{1-1}
\hline
{Avg.} & {35.93} & {0.38} & {73.40}  & {38.07}    & {0.35}& {33.38}  
\\ \bottomrule
\end{tabular}
}
\caption{The average metrics on DTU-Robust scan106 before and after MV-Prune, including PSNR, CD, and Gaussian file size.}
    \label{tab: mv-prune}
\end{table}

\section*{Mask Analysis} 
\label{mask}
\begin{figure}[h]
    \centering
    \begin{subfigure}{0.3\linewidth}
        \centering
        \includegraphics[width=\linewidth]{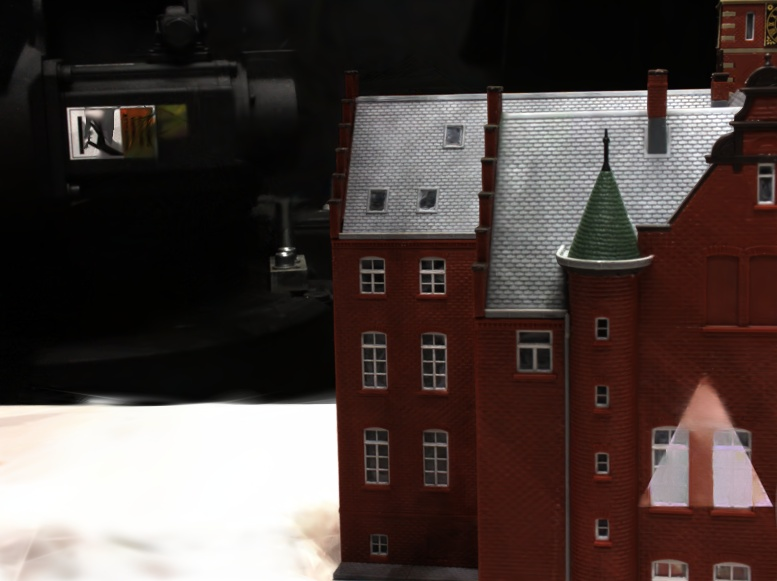}
        \caption{w/o mask}
        \label{fig:mask_a}
    \end{subfigure}
    \begin{subfigure}{0.3\linewidth}
        \centering
        \includegraphics[width=\linewidth]{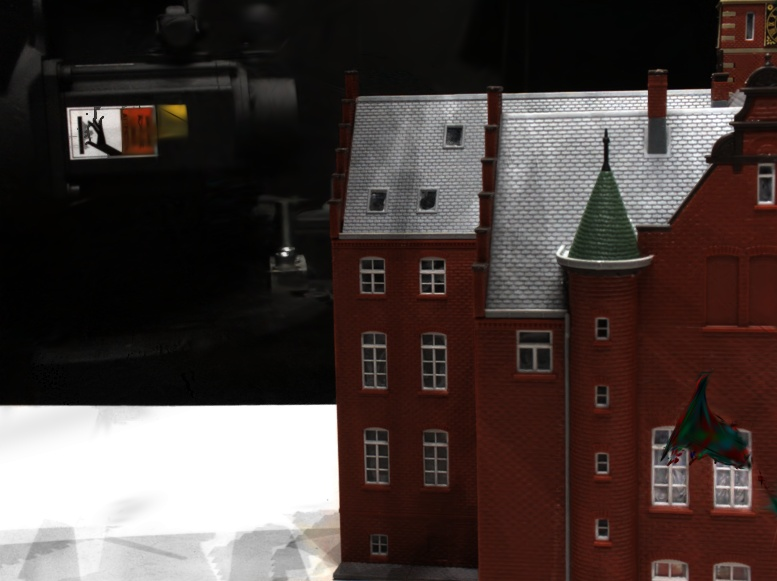} 
        \caption{$Mask_{mv}$}
        \label{fig:mask_b}
    \end{subfigure}
    \begin{subfigure}{0.3\linewidth}
        \centering
        \includegraphics[width=\linewidth]{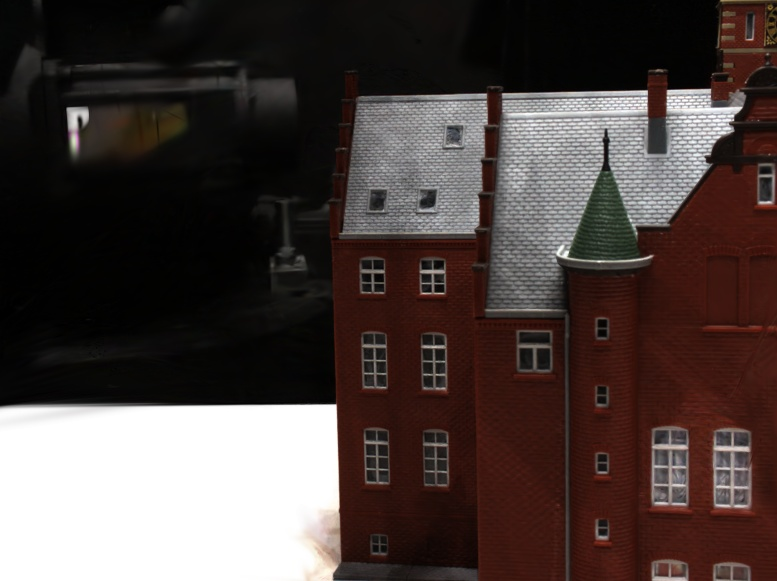} 
        \caption{$Mask_{sam}$}
        \label{fig:mask_c}
    \end{subfigure}
    \caption{Comparison of different mask strategies.}
    \label{fig:mask_comparison}
\end{figure}

We show the analysis of the effectiveness of our proposed masking strategies in Fig.~\ref{fig:mask_comparison}. The baseline `w/o mask' renders the scene with a noticeable artifact - the distractor as shown in Fig.~\ref{fig:mask_a}, and `$Mask_{mv}$' improves the geometry consistency and renders the details behind the distractor with small artifacts, as shown in Fig.~\ref{fig:mask_b}. The `$Mask_{sam}$' achieves further enhancement by combining the results from SAM, and provides clear and accurate rendering results, as shown in Fig~\ref{fig:mask_c}. The analysis proves that our integrated masking approach effectively mitigates visual artifacts caused by the distractor and improves the overall rendering quality.

\begin{figure*}[h]
    \centering
    \includegraphics[width=\linewidth]{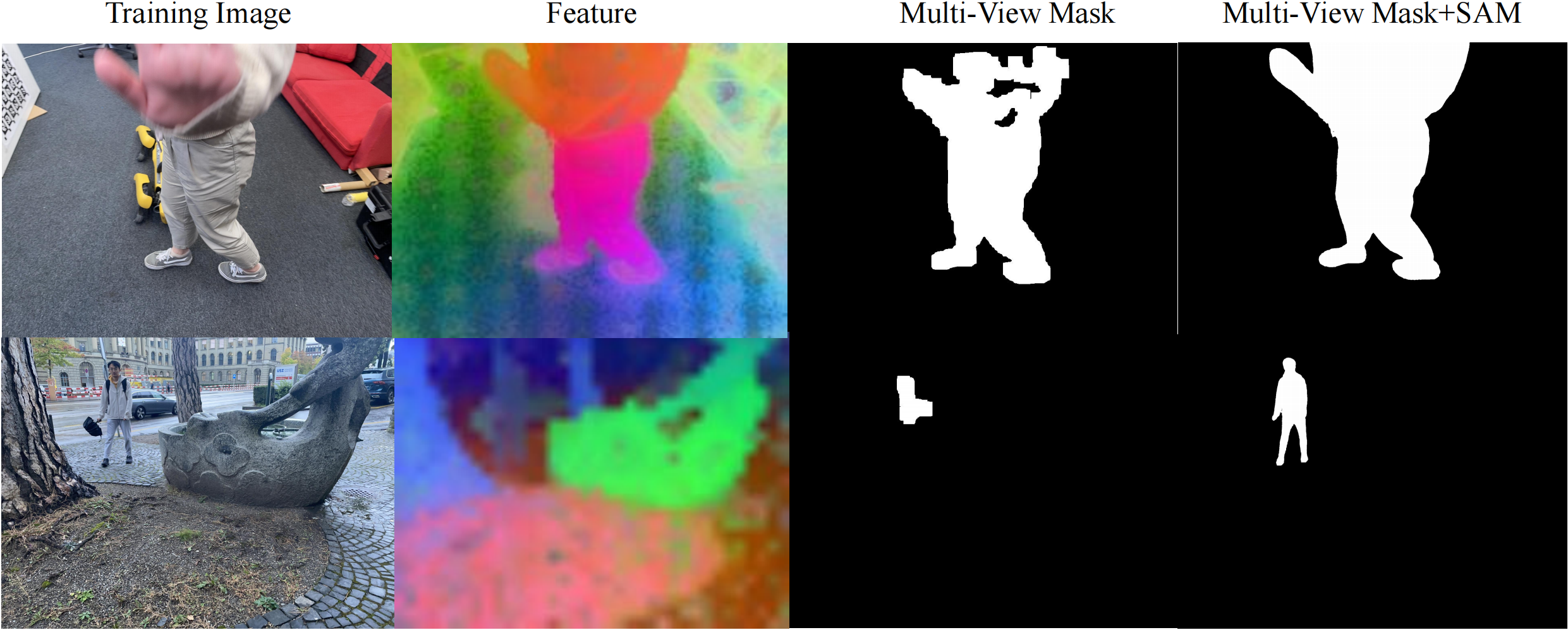}
    \caption{Disturbed images, corresponding semantic features, and multi-view mask results on NeRF onthego Dataset}
    \label{fig:Mask}
\end{figure*}

As shown in Fig. \ref{fig:Mask}, the semantic features of disturbed images can visually distinguish the distractors. By comparing features from multiple views, it is possible to obtain hints of disturbances before training. SAM is then used to refine the distractor masks. These optimized masks can enhance the model's accuracy in handling complex scenes. This approach not only allows us to identify potential disturbances before training but also enables dynamic adjustments during the training process, improving overall performance and robustness. Experiments on the DTU-Robust dataset demonstrate that this method achieves significant improvements across various scenarios.

\section{Ablation Study and Runtime}
\label{ablation}

\begin{table}[h]

\centering
\resizebox{\linewidth}{!}{
\begin{tabular}{cccc}
\cline{1-4}
\multicolumn{1}{l|}{}     & delta\_near=0.3 & delta\_near=0.5 & delta\_near=0.8       \\ \cline{1-4}
\multicolumn{1}{c|}{PSNR↑} & 32.56                & 33.06               & 33.10                    \\
\multicolumn{1}{c|}{CD↓}   & 0.54               & 0.34                &  0.60                  \\ \cline{1-4}
\end{tabular}
}

\caption{Ablation Study in the setting of delta\_near. }
\label{table:ablation3}
\end{table}

\begin{table}[h]
\centering
\resizebox{\linewidth}{!}{
\begin{tabular}{c|cccccc}
\cline{1-7}
            & \multicolumn{1}{l}{CD↓} & \multicolumn{1}{l}{PSNR↑} & Time/min & Memory/GB & Size/MB & Points \\ \hline
NeRF on-the-go & 1.05                      &  17.26                    & 240         & 60.0          & 103.4       & -      \\ \cline{1-1}
SpotLessSplats         & 0.90                      & 27.27                    & 9.95     & 3.06      & 80.97    & 400k       \\ \hline
PGSR        & 0.53                   & 31.31                    & 33.3    &  4.11      & 89.8    & 379k        \\ \cline{1-1}
2DGS        & 0.57                   & 30.96                    & 17.1     & 3.85      & 2.7     & 11k   \\ \hline
MVGSR w/o MV-Prune      & 0.37                   & 32.33                    & 39.8     & 4.45      &133.0   & 560k    \\ MVGSR        & 0.34                   & 33.06                  & 39.9     & 4.45      &59.9   & 260k    \\ \hline

\end{tabular}
}

\resizebox{\linewidth}{!}{
\begin{tabular}{c|cccccc|c}
     MVGSR& train\_7k & render & extract\_feature & mask\_mv & mask\_sam & train   & total    \\ \hline
Time & 123.2s    & 19.8s  & 41.8s            & 91.7s    & 89.9s     & 2029.0s & 39.93min \\ \hline
\end{tabular}
}
\caption{The Runtime metrics on DTU-Robust scan24}
\label{table:runtime2_y}
\end{table}

To demonstrate the contribution of each component, ablation experiments were conducted on DTU-Robust scan24 in this section. Also, as shown in Tab. \ref{table:runtime2_y}, we compare the runtime and resource consumption of different algorithms on scan24 of the DTU-Robust dataset. Our approach achieves optimal results with lightweight resource utilization.

\begin{figure*}[h]
    \centering
    \includegraphics[width=\linewidth]{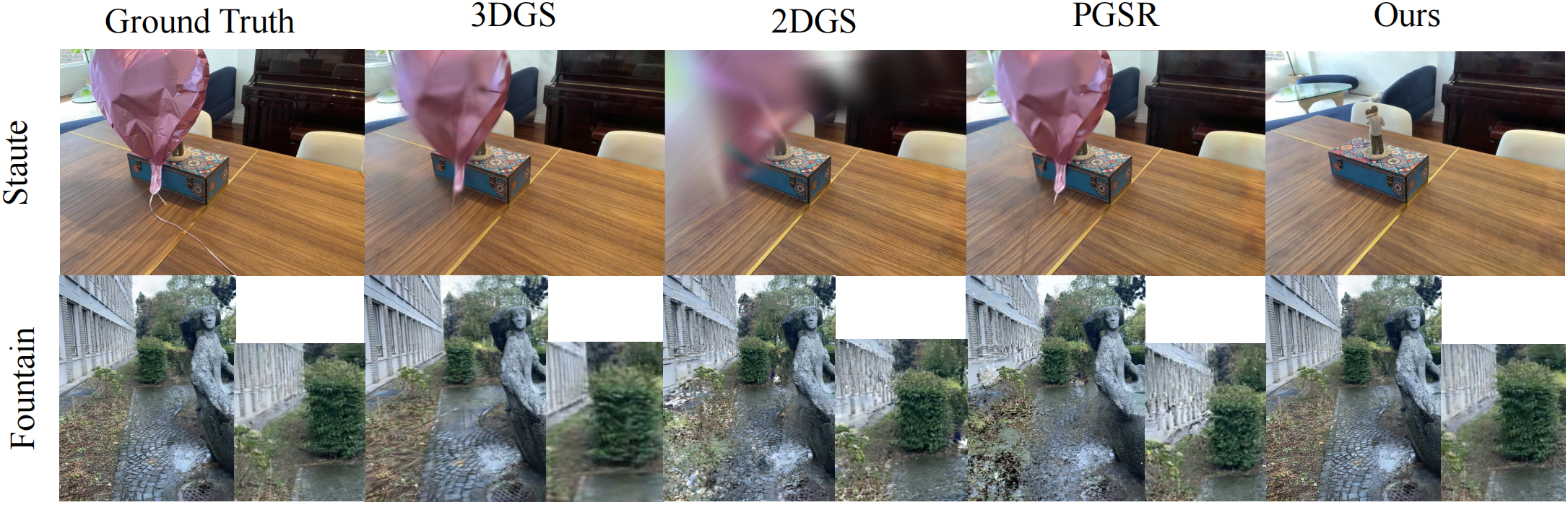} 
    \caption{Visualization results of the t\_balloon\_statue on the dataset provided by RobustNeRF and the fountain on the NeRF-On-the-GO }
    \label{fig:robustnerf2}
\end{figure*}

\section{More Results on DTU-Robust Dataset}
\label{dtu}
\subsection{Dataset Analysis}
The DTU-Robust dataset introduces distractors with random positions, colors, and shapes to a given proportion (rate) of DTU training images. The relationship between the number of disturbed images and the number of disturbed pixels is shown in Fig. \ref{fig:dtu-robust}. Specifically, at a \textit{rate 0.3}, the distractor regions occupy 9.45\% of the image, while this increases to 28.34\% at \textit{rate 1.0}. Therefore, from the perspective of the occupied pixels in the distractor areas, the challenges increase from \textit{rate 0.3} to \textit{rate 1.0} for the same sequence.

The DTU-Robust dataset effectively demonstrates the impact of occlusions, highlighting reconstruction errors and novel view rendering errors in the presence of these distractions.

\begin{table*}
\centering
\resizebox{0.8\textwidth}{!}{
\begin{tabular}{cc|c|c|c|c|c|c|c}
\toprule
\multicolumn{2}{c|}{scan} & Truck & Caterpillar & Barn & Meetingroom & Ignatius & Courthouse & Avg. \\ \cline{1-9}
\multirow{4}{*}{F1$\uparrow$} 
&PGSR~\cite{chen2024pgsr} 
& 0.57 & 0.37 & {0.57} & 0.30 & 0.64 & {0.18} & 0.44 \\ \cline{2-9}
&SLS~\cite{sabourgoli2024spotlesssplats} 
& 0.48 & 0.30 & 0.43 & 0.24 & 0.57 & 0.15 & 0.36 \\ \cline{2-9}
&NeRFonthego~\cite{Ren2024NeRFonthego} 
& 0.37 & 0.22 & 0.28 & 0.17 & 0.49 & 0.11 & 0.27  \\ \cline{2-9}
&MVGSR 
& {0.60} & {0.42} & {0.59} & {0.34} & {0.73} & 0.18 & {0.48} \\

\cline{1-9}
\multirow{4}{*}{PSNR$\uparrow$} 
&PGSR~\cite{chen2024pgsr} 
& {21.67} & {20.35} & 23.21 & 21.94 & 18.79 & 20.74 &20.95 \\ \cline{2-9}
&SLS~\cite{sabourgoli2024spotlesssplats} 
 & 20.66 & 21.14 & 21.77 & 21.88 & 20.67 & 20.44 & 21.09 \\ \cline{2-9}
&NeRFonthego~\cite{Ren2024NeRFonthego} 
& 19.33 & 19.10 & 23.46 & 21.95 & 17.36 & 20.97 & 20.36\\ \cline{2-9}
&MVGSR 
& 21.64 & 19.79 & {23.46} & {21.96} & 18.95 & 20.97 & 21.13 \\ 

\bottomrule
\end{tabular}
}
\caption{Quantitative results of reconstruction performance on the TnT-Robust dataset.}
\label{table_tnt2}
\end{table*}

\subsection{Reconstruction}

For surface quality evaluation, we use two other metrics, in addition to the bidirectional Chamfer distance \textcolor{black}{CD} used in the main paper, to assess the similarity between point clouds. 
We denote the Chamfer distance from the destination to the source as \textcolor{black}{d2s}, and from the source to the destination as \textcolor{black}{s2d}.

We compare the reconstruction performance of our proposed method to PGSR~\cite{chen2024pgsr} and 2DGS~\cite{huang20242d} on the DTU-Robust dataset, across different scans and varying occlusion rates. As shown in Tab.~\ref{table_2_recon_CD2}, our method achieves superior and comparable performance under various evaluation protocols (CD, d2s, and s2d metrics) for the additional scans in the dataset.

Notably, our method outperforms both PGSR and 2DGS significantly in all scans. Specifically for \textit{scan 24}, at an occlusion rate of 0.8, our method achieves a CD score that is more than twice as good as PGSR, improving from 0.83 to 0.34. Compared to 2DGS and PGSR, our method exhibits greater robustness across all scenes, with fewer performance fluctuations. We achieve comparable performance to PGSR for all other settings.

On average across all settings, our method outperforms 2DGS and PGSR by 76.4\% and 60.3\%, with an average CD score of 0.42. These results demonstrate that our method maintains robust reconstruction quality across varying occlusion rates and different scenes.
\begin{table*}
\centering
\resizebox{\textwidth}{!}{
\begin{tabular}{cc|cccc|cccc|cccc|cccc}
\\\toprule
\multicolumn{2}{l|}{scan}  
& \multicolumn{4}{c|}{24}  & \multicolumn{4}{c|}{37} & \multicolumn{4}{c|}{40} & \multicolumn{4}{c}{55} \\ \cline{1-18}

\multicolumn{2}{l|}{rate}  
& \multicolumn{1}{c|}{0.3} & \multicolumn{1}{c|}{0.5} & \multicolumn{1}{c|}{0.8} & 1.0  & \multicolumn{1}{c|}{0.3} & \multicolumn{1}{c|}{0.5} & \multicolumn{1}{c|}{0.8} & 1.0 & \multicolumn{1}{c|}{0.3} & \multicolumn{1}{c|}{0.5} & \multicolumn{1}{c|}{0.8} & 1.0 & \multicolumn{1}{c|}{0.3} & \multicolumn{1}{c|}{0.5} & \multicolumn{1}{c|}{0.8} & 1.0 \\ \hline

\multirow{3}{*}{CD$\downarrow$} 
&PGSR~\cite{chen2024pgsr}  
       &0.39&0.37&0.83&0.53 
       &0.66&0.96&1.22&1.16 
       &0.44&0.51&0.46&0.62 
       &0.36&0.36&0.35&0.36\\ \cline{2-1}
&2DGS~\cite{huang20242d}  
        &0.50 &0.52 &0.48 &0.57
        & 0.83 & 0.85 &0.85 &0.92
        & 0.34 &0.69 &0.92 &0.43
        & 0.38 & 0.39 &0.45 & 0.56 \\ \cline{2-1}
&MVGSR 
        &0.34&0.34&0.34&0.33
        &0.50&0.50&0.51&0.52
        &0.30&0.30&0.30&0.31
        &0.30&0.30&0.30&0.30
        
\\ \hline
\multirow{3}{*}{d2s$\downarrow$} 
&PGSR~\cite{chen2024pgsr}  
        &0.38&0.36&1.23&0.70 
        &0.78&1.39&1.91&1.79
        &0.51&0.66&0.54&0.87
        &0.33&0.32&0.29&0.31 \\ \cline{2-1}
&2DGS~\cite{huang20242d}  
        &0.50 &0.54 &0.47 &0.58
        & 0.99&1.02 &1.03&1.18
        &  0.35&0.48 &0.59 &0.53
        & 0.35 & 0.36 &0.46 & 0.65 \\ \cline{2-1}
&MVGSR 
      &0.31&0.32&0.32&0.31 
       &0.51&0.52&0.54&0.55
       &0.30&0.30&0.30&0.32 
       &0.30&0.29&0.3&0.30 \\ \hline 

\multirow{3}{*}{s2d$\downarrow$} 
&PGSR~\cite{chen2024pgsr} 
        &0.40&0.37&0.44&0.37 
        &0.54&0.54&0.54&0.53
        &0.37&0.37&0.38&0.38
        &0.38&0.40&0.41&0.40  \\ \cline{2-1}
&2DGS~\cite{huang20242d}  
        &0.50 &0.51 &0.49 &0.57
        & 0.67 & 0.67 &0.66 &0.66
        &  0.33 &1.24 &0.90 &0.34
        & 0.41 & 0.43 &0.44 & 0.48 \\ \cline{2-1}
&MVGSR 
       &0.36&0.36&0.36&0.35 
       &0.48&0.48&0.48&0.49
       &0.30&0.30&0.30&0.31
       &0.31&0.30&0.31&0.31  \\ \midrule  \toprule

\multicolumn{2}{l|}{scan}  
& \multicolumn{4}{c|}{63}  & \multicolumn{4}{c|}{65} & \multicolumn{4}{c|}{69} & \multicolumn{4}{c}{83} \\ \cline{1-18}

\multicolumn{2}{l|}{rate}  
& \multicolumn{1}{c|}{0.3} & \multicolumn{1}{c|}{0.5} & \multicolumn{1}{c|}{0.8} & 1.0  & \multicolumn{1}{c|}{0.3} & \multicolumn{1}{c|}{0.5} & \multicolumn{1}{c|}{0.8} & 1.0 & \multicolumn{1}{c|}{0.3} & \multicolumn{1}{c|}{0.5} & \multicolumn{1}{c|}{0.8} & 1.0 & \multicolumn{1}{c|}{0.3} & \multicolumn{1}{c|}{0.5} & \multicolumn{1}{c|}{0.8} & 1.0 \\ \hline

\multirow{3}{*}{CD$\downarrow$} 
&PGSR~\cite{chen2024pgsr}  
        &0.75&1.17&1.17&1.46 
        &0.60&0.64&0.61&0.59
        &0.48&0.49&0.49&0.51
        &0.82&0.85&1.02&0.96\\ \cline{2-1}
&2DGS~\cite{huang20242d}  
        &1.07 &1.02 &1.16 &1.22
        & 0.97 & 1.09 &0.92 &1.00
        & 0.79 & 0.81 &0.85& 0.81
        &  1.32& 1.33 &1.39 &1.42
        \\ \cline{2-1}
&MVGSR 
        &0.43&0.43&0.44&0.44
        &0.50&0.53&0.52&0.51 
        &0.45&0.45&0.45&0.45
        &0.62&0.62&0.62&0.63          
\\ \hline
\multirow{3}{*}{d2s$\downarrow$} 
&PGSR~\cite{chen2024pgsr}  
        &1.09&1.93&1.91&2.52 
        &0.58&0.58&0.55&0.57
        &0.50&0.50&0.51&0.51
        &0.55&0.84&0.90&1.07 \\ \cline{2-1}
&2DGS~\cite{huang20242d}  
        &1.18 &1.13 &1.25 &1.30
        & 0.89 & 1.08 &0.91 &1.02
        & 0.79 &0.77 &0.84 &0.79
        & 1.00& 0.95 &1.00 & 1.04 \\ \cline{2-1}
&MVGSR 
        &0.52&0.51&0.53&0.52
        &0.52&0.55&0.55&0.54
        &0.47&0.46&0.47&0.47 
        &0.56&0.55&0.56&0.56 \\ \hline 

\multirow{3}{*}{s2d$\downarrow$} 
&PGSR~\cite{chen2024pgsr} 
        &0.41&0.42&0.43&0.41 
        &0.62&0.70&0.67&0.61
        &0.47&0.47&0.48&0.51
        &1.09&0.86&1.14&0.84 \\ \cline{2-1}
&2DGS~\cite{huang20242d}  
        &0.96 &0.91&1.07 &1.14
        & 1.06&1.11 &0.93&0.97
        & 0.80 &0.84&0.85&0.84
        & 1.63 & 1.71 &1.78 & 1.81 \\ \cline{2-1}
&MVGSR  
        &0.34&0.35&0.36&0.35 
        &0.47&0.50&0.50&0.48
        &0.43&0.43&0.44&0.44
        &0.68&0.69&0.69&0.71        
        \\ \midrule \toprule
        
\multicolumn{2}{l|}{scan}  
& \multicolumn{4}{c|}{97}  & \multicolumn{4}{c|}{105} & \multicolumn{4}{c|}{106} & \multicolumn{4}{c}{110} \\ \cline{1-18}

\multicolumn{2}{l|}{rate}  
& \multicolumn{1}{c|}{0.3} & \multicolumn{1}{c|}{0.5} & \multicolumn{1}{c|}{0.8} & 1.0  & \multicolumn{1}{c|}{0.3} & \multicolumn{1}{c|}{0.5} & \multicolumn{1}{c|}{0.8} & 1.0 & \multicolumn{1}{c|}{0.3} & \multicolumn{1}{c|}{0.5} & \multicolumn{1}{c|}{0.8} & 1.0 & \multicolumn{1}{c|}{0.3} & \multicolumn{1}{c|}{0.5} & \multicolumn{1}{c|}{0.8} & 1.0 \\ \hline

\multirow{3}{*}{CD$\downarrow$} 
&PGSR~\cite{chen2024pgsr}  
        &0.63&0.63&0.62&0.63 
        &0.58&0.59&0.61&0.73 
        &0.47&0.47&0.47&0.49
        &0.51&0.46&0.45&0.44 \\ \cline{2-1}
&2DGS~\cite{huang20242d}  
        &1.22 &1.19 &1.22&1.13
        & 0.68 & 0.68&0.69 &0.69
        &  0.70 &0.69 &0.70 &0.73
        & 0.70 & 0.69&0.70 & 0.73 \\ \cline{2-1}
&MVGSR 
       &0.59&0.59&0.60&0.59
       &0.55&0.56&0.56&0.56
       &0.35&0.35&0.35&0.37
       &0.38&0.37&0.36&0.38 
        
\\ \hline
\multirow{3}{*}{d2s$\downarrow$} 
&PGSR~\cite{chen2024pgsr}  
        &0.63&0.64&0.61&0.65
        &0.49&0.50&0.54&0.79
        &0.36&0.37&0.36&0.39
        &0.63&0.55&0.52&0.51  \\ \cline{2-1}
&2DGS~\cite{huang20242d}  
        &1.09 &1.01&1.08 &0.96
        & 0.60 & 0.59 &0.61 &0.60
        &  0.50 &0.48 &0.50 &0.58
        & 1.44 & 1.60 &1.53 & 1.63 \\ \cline{2-1}
&MVGSR 
       &0.55&0.57&0.57&0.56 
       &0.45&0.46&0.46&0.46
       &0.34&0.33&0.33&0.37
       &0.39&0.38&0.36&0.39 \\ \hline 

\multirow{3}{*}{s2d$\downarrow$} 
&PGSR~\cite{chen2024pgsr} 
        &0.62&0.62&0.62&0.61
        &0.68&0.68&0.69&0.68
        &0.58&0.57&0.57&0.58
        &0.38&0.38&0.38&0.38  \\ \cline{2-1}
&2DGS~\cite{huang20242d}  
        &1.35 &1.36 &1.36&1.30
        &0.77 & 0.78 &0.77 &0.78
        &  0.90 &0.90 &0.90 &0.88
        & 1.23 &1.56 &1.60 & 1.57\\ \cline{2-1}
&MVGSR  
        &0.62&0.62&0.63&0.62
        &0.65&0.66&0.66&0.66
        &0.36&0.36&0.36&0.37
        &0.38&0.37&0.36&0.37 \\ \midrule \toprule
        
\multicolumn{2}{l|}{scan}  
& \multicolumn{4}{c|}{114}  & \multicolumn{4}{c|}{118} & \multicolumn{4}{c|}{122} & \multicolumn{4}{c}{Avg.} \\ \cline{1-18}

\multicolumn{2}{l|}{rate}  
& \multicolumn{1}{c|}{0.3} & \multicolumn{1}{c|}{0.5} & \multicolumn{1}{c|}{0.8} & 1.0  & \multicolumn{1}{c|}{0.3} & \multicolumn{1}{c|}{0.5} & \multicolumn{1}{c|}{0.8} & 1.0 & \multicolumn{1}{c|}{0.3} & \multicolumn{1}{c|}{0.5} & \multicolumn{1}{c|}{0.8} & 1.0 & \multicolumn{1}{c|}{0.3} & \multicolumn{1}{c|}{0.5} & \multicolumn{1}{c|}{0.8} & 1.0 \\ \cline{1-18}

\multirow{3}{*}{CD$\downarrow$} 
&PGSR~\cite{chen2024pgsr}  
        &0.32&0.31&0.32&0.36 
        &0.37&0.37&0.44&0.38
        &0.35&0.34&0.37&0.37
        &0.52&0.57 &0.63 &0.64\\ \cline{2-1}
&2DGS~\cite{huang20242d}  
        &0.38 &0.39 &0.48 &0.39
        & 0.68& 0.71 &0.69 &0.71
        & 0.51 &0.54 &0.56 &0.64
        &0.74&0.77 &0.80 &0.80\\ \cline{2-1}
&MVGSR 
        &0.29&0.29&0.30&0.29 
        &0.34&0.34&0.34&0.35 
        &0.34&0.34&0.35&0.35
        &0.42&0.43&0.44&0.40\\ \cline{1-18}

\multirow{3}{*}{d2s$\downarrow$} 
&PGSR~\cite{chen2024pgsr}  
        &0.33&0.33&0.33&0.33 
        &0.40&0.40&0.42&0.43
        &0.38&0.38&0.38&0.41 
        &0.53&0.65 &0.73 &0.79 \\ \cline{2-1}
&2DGS~\cite{huang20242d}  
        &0.33 &0.35 &0.43 &0.34
        &0.55 &0.57&0.60&0.60
        &0.49 &0.50 &0.56 &0.61 
        &0.74&0.76 &0.79 &0.83\\ \cline{2-1}
&MVGSR 
        &0.26&0.26&0.26&0.26 
        &0.32&0.34&0.33&0.34
        &0.32&0.32&0.33&0.33
        &0.44&0.46&0.48&0.43\\ \cline{1-18}

\multirow{3}{*}{s2d$\downarrow$} 
&PGSR~\cite{chen2024pgsr} 
        &0.32&0.31&0.32&0.36
        &0.37&0.37&0.44&0.38
        &0.35&0.34&0.37&0.37
        &0.51&0.50 &0.53 &0.49 \\ \cline{2-1}
&2DGS~\cite{huang20242d}  
        &0.44 &0.43 &0.53 &0.44
        &0.81&0.86 &0.78 &0.81
        &0.54&0.59 &0.55 &0.67
        &0.83&0.93 &0.91 &0.88\\ \cline{2-1}
&MVGSR  
        &0.33&0.33&0.33&0.32
        &0.35&0.35&0.35&0.37 
        &0.36&0.36&0.36&0.37
        &0.43&0.45&0.47 &0.44 \\ \bottomrule

\end{tabular}
}
\vspace{-2mm}
\caption{Quantitative results of reconstruction performance on the DTU-Robust dataset. }
\vspace{-3mm}
\label{table_2_recon_CD2}
\end{table*}

\subsection{Rendering}
\label{subsec:app_rendering}

To evaluate the novel view rendering performance of our proposed method, we compare with PGSR~\cite{chen2024pgsr}, 2DGS~\cite{huang20242d} and, additionally, 3DGS ~\cite{kerbl20233d} for the novel view rendering task on the same DTU-Robust dataset. As shown in Tab. \ref{table_1_render_psnr}, our method is largely able to outperform the above competing methods. When comparing PSNR, our method achieves best performance on average under all settings for scenes that are evaluated on. 

Similar to the task of reconstruction, our method exhibits greater robustness across all scenes, with fewer performance fluctuations compared to other competing methods. 
On average, we outperform 3DGS most significantly by close to 6.82 dB with a PSNR of 34.42 dB while outperforming other methods by a smaller margin. These results demonstrate that our method can improve the quality of rendered images across varying occlusion rates and different scenes.
\begin{table*}
\centering
\resizebox{\textwidth}{!}{
\begin{tabular}{cc|cccc|cccc|cccc|cccc}
\hline
\multicolumn{2}{l|}{scan} & \multicolumn{4}{c|}{24} & \multicolumn{4}{c|}{37} & \multicolumn{4}{c|}{40} & \multicolumn{4}{c}{55} \\ \cline{1-18}
\multicolumn{2}{l|}{rate} & \multicolumn{1}{c|}{0.3} & \multicolumn{1}{c|}{0.5} & \multicolumn{1}{c|}{0.8} & 1.0 & \multicolumn{1}{c|}{0.3} & \multicolumn{1}{c|}{0.5} & \multicolumn{1}{c|}{0.8} & 1.0 & \multicolumn{1}{c|}{0.3} & \multicolumn{1}{c|}{0.5} & \multicolumn{1}{c|}{0.8} & 1.0 & \multicolumn{1}{c|}{0.3} & \multicolumn{1}{c|}{0.5} & \multicolumn{1}{c|}{0.8} & 1.0 \\ \hline
\multirow{3}{*}{PSNR$\uparrow$} 
& 3DGS~\cite{kerbl20233d} 
        & 26.21 & 25.46 & 25.26 & 24.74 
        &24.56&24.92&23.84&22.25 
        &22.86&23.55&22.84&22.05
        &30.34&29.35&28.07&25.99 \\
& PGSR~\cite{chen2024pgsr} 
        &31.86&31.49&30.68&31.31
        &27.51&27.07&26.56&25.73 
        &30.83&30.56&29.48&28.66
        &32.95&32.11&31.73&30.44 \\
& 2DGS~\cite{huang20242d} 
        & 33.47& 28.94 & 27.87 & 30.96 
        &28.96& 29.07&28.86&27.94
        &29.94&27.89& 32.77&21.02
        &21.02&23.92&30.44& 33.22 \\
& MVGSR 
        &33.61 &32.95 &32.41 &32.28 
        &30.05 &29.89 &29.27 &27.68
        &33.43 &33.03 &32.3 &31.33
        &35.19 &35.33 &33.99 &32.66 \\ \hline
\multirow{3}{*}{SSIM$\uparrow$} 
& 3DGS~\cite{kerbl20233d} 
        &0.926&0.921&0.919&0.916
        &0.955&0.959&0.952&0.939 
        &0.935&0.943&0.942&0.927
        &0.963&0.961&0.953&0.944
        \\
& PGSR~\cite{chen2024pgsr}
        &0.946&0.944&0.941&0.943
        &0.942&0.940&0.936&0.931
        &0.941&0.940&0.935&0.932
        &0.923&0.920&0.919&0.913  \\
& 2DGS~\cite{huang20242d} 
        & 0.958 & \underline{0.909} & 0.906 & 0.950
        &0.950& 0.958&0.956&0.951
        &0.951&0.944& 0.952&0.657 
        &0.657&0.882&0.940&0.927 \\
& MVGSR 
        &0.956 &0.954 &0.952 &0.951 
        &0.963 &0.962 &0.959 &0.953
        &0.954 &0.953 &0.95 &0.945
        &0.981 &0.982 &0.979 &0.974  \\ \hline
\multirow{3}{*}{LPIPS$\downarrow$} 
& 3DGS~\cite{kerbl20233d} 
        &0.086&0.094&0.095&0.098 
        &0.050&0.046&0.053&0.068
        &0.098&0.093&0.095&0.115
        &0.060&0.063&0.072&0.082  \\
& PGSR~\cite{chen2024pgsr} 
        &0.068&0.073&0.076&0.075 
        &0.105&0.107&0.113&0.119 
        &0.118&0.121&0.128&0.133
        &0.138&0.144&0.146&0.153  \\
& 2DGS~\cite{huang20242d} 
        & 0.073 & 0.140 & 0.146 & 0.082 
        &0.082&0.096&0.098&0.103 
        &0.103&0.112&0.111&0.448 
        &0.448&0.189&0.128&0.168\\
& MVGSR 
        &0.042 &0.046 &0.048 &0.05 
        &0.038 &0.04 &0.042 &0.05
        &0.081 &0.084 &0.089 &0.095
        &0.031 &0.029 &0.034 &0.042\\ \midrule
        \toprule

\multicolumn{2}{l|}{scan} & \multicolumn{4}{c|}{63} & \multicolumn{4}{c|}{65} & \multicolumn{4}{c|}{69} & \multicolumn{4}{c}{83} \\ \cline{1-18}
\multicolumn{2}{l|}{rate} & \multicolumn{1}{c|}{0.3} & \multicolumn{1}{c|}{0.5} & \multicolumn{1}{c|}{0.8} & 1.0 & \multicolumn{1}{c|}{0.3} & \multicolumn{1}{c|}{0.5} & \multicolumn{1}{c|}{0.8} & 1.0 & \multicolumn{1}{c|}{0.3} & \multicolumn{1}{c|}{0.5} & \multicolumn{1}{c|}{0.8} & 1.0 & \multicolumn{1}{c|}{0.3} & \multicolumn{1}{c|}{0.5} & \multicolumn{1}{c|}{0.8} & 1.0 \\ \hline
\multirow{3}{*}{PSNR$\uparrow$} 
& 3DGS~\cite{kerbl20233d} 
        &26.91&26.78&24.56&23.20 
        &29.32&27.69&25.53&24.57
        &28.64&27.29&25.54&24.38 
        &25.96&25.70&25.06&23.39 \\
& PGSR~\cite{chen2024pgsr} 
        &33.33&32.77&32.54&32.00 
        &32.94&31.88&30.44&29.80
        &31.79&30.99&30.96&30.02 
        &32.79&32.05&32.00&31.22 \\
& 2DGS~\cite{huang20242d} 
        &33.22&32.99&31.47&30.37
        &35.38&33.86&32.52&30.82 
        &31.82&33.57&31.94&31.04 
        &31.04&29.93&32.34&32.16 \\
& MVGSR 
        &38.68 &38.06 &36.37 &34.94 
        &36.07 &34.42 &33.72 &32.24
        &33.68 &33.27 &33 &31.47
        &41.78 &41.47 &40.9 &39.44  \\ \hline
\multirow{3}{*}{SSIM$\uparrow$} 
& 3DGS~\cite{kerbl20233d} 
        &0.968&0.965&0.957&0.947 
        &0.968&0.961&0.953&0.951 
        &0.947&0.939&0.919&0.925
        &0.965&0.959&0.958&0.947\\
& PGSR~\cite{chen2024pgsr}
        &0.950&0.948&0.946&0.943
        &0.917&0.915&0.910&0.911
        &0.917&0.914&0.912&0.908
        &0.909&0.905&0.905&0.902 \\
& 2DGS~\cite{huang20242d} 
        &0.927&0.925&0.921&0.914
        &0.965&0.962&0.959&0.955
        &0.955&0.930&0.927&0.924 
        &0.924&0.922&0.929&0.930 \\
& MVGSR 
        &0.978 &0.977 &0.974 &0.972 
        &0.979 &0.977 &0.976 &0.973
        &0.965 &0.965 &0.963 &0.958
        &0.987 &0.987 &0.986 &0.984 \\ \hline
\multirow{3}{*}{LPIPS$\downarrow$} 
& 3DGS~\cite{kerbl20233d} 
        &0.038&0.045&0.056&0.069 
        &0.046&0.055&0.068&0.069
        &0.072&0.084&0.103&0.100
        &0.048&0.055&0.057&0.068
        \\
& PGSR~\cite{chen2024pgsr} 
        &0.106&0.110&0.112&0.116
        &0.163&0.165&0.174&0.171
        &0.145&0.149&0.151&0.158
        &0.216&0.229&0.228&0.234 \\
& 2DGS~\cite{huang20242d} 
        &0.168&0.170&0.177&0.188 
        &0.101&0.108&0.111&0.119  
        &0.119&0.172&0.176&0.181  
       &0.181&0.185&0.152&0.152  \\
& MVGSR 
        &0.024 &0.026 &0.03 &0.034 
        &0.039 &0.042 &0.045 &0.048
        &0.057 &0.058 &0.062 &0.067
        &0.022 &0.024 &0.025 &0.027 \\ \midrule 
        \toprule

\multicolumn{2}{l|}{scan} & \multicolumn{4}{c|}{97} & \multicolumn{4}{c|}{105} & \multicolumn{4}{c|}{106} & \multicolumn{4}{c}{110} \\ \cline{1-18}
\multicolumn{2}{l|}{rate} & \multicolumn{1}{c|}{0.3} & \multicolumn{1}{c|}{0.5} & \multicolumn{1}{c|}{0.8} & 1.0 & \multicolumn{1}{c|}{0.3} & \multicolumn{1}{c|}{0.5} & \multicolumn{1}{c|}{0.8} & 1.0 & \multicolumn{1}{c|}{0.3} & \multicolumn{1}{c|}{0.5} & \multicolumn{1}{c|}{0.8} & 1.0 & \multicolumn{1}{c|}{0.3} & \multicolumn{1}{c|}{0.5} & \multicolumn{1}{c|}{0.8} & 1.0 \\ \hline
\multirow{3}{*}{PSNR$\uparrow$} 
& 3DGS~\cite{kerbl20233d} 
        &23.35&25.67&24.89&24.19 
        &24.21&24.36&23.33&22.89
        &33.98&34.21&33.24&31.19
        &35.31&34.57&34.50&33.98 \\
& PGSR~\cite{chen2024pgsr} 
        &30.80&30.89&30.30&29.49
        &33.63&33.09&32.38&31.92
        &35.40&35.43&35.05&33.86 
        &34.35&32.71&33.96&33.28 
        \\
& 2DGS~\cite{huang20242d} 
        &32.16&31.74&30.59&33.58 
        &33.58&32.99&19.52&18.21
        &31.01&30.68&30.33&29.84
        &29.84&33.80&33.42&32.31\\
& MVGSR 
        &35.21 &34.99 &34.44 &32.83 
        &39.3 &38.66 &37.74 &36.38
        &38.29 &38.59 &37.99 &36.98
        &36.35 &35.93 &35.61 &35.34 \\ \hline
\multirow{3}{*}{SSIM$\uparrow$} 
& 3DGS~\cite{kerbl20233d} 
        &0.936&0.953&0.942&0.940 
        &0.952&0.952&0.945&0.941
        &0.959&0.959&0.957&0.950
        &0.961&0.959&0.958&0.957\\
& PGSR~\cite{chen2024pgsr}
        &0.910&0.909&0.909&0.905
        &0.916&0.914&0.912&0.909
        &0.933&0.933&0.933&0.928
        &0.923&0.917&0.919&0.918  \\
& 2DGS~\cite{huang20242d} 
        &0.930&0.927&0.922&0.918 
        &0.918&0.916&0.789&0.766 
        &0.919&0.918&0.916&0.914
        &0.914&0.925&0.923&0.920\\
& MVGSR 
        &0.974 &0.973 &0.973 &0.968 
        &0.977 &0.976 &0.974 &0.972
        &0.977 &0.978 &0.977 &0.974
        &0.974 &0.973 &0.971 &0.971 \\ \hline
\multirow{3}{*}{LPIPS$\downarrow$} 
& 3DGS~\cite{kerbl20233d} 
        &0.072&0.055&0.068&0.069
        &0.061&0.063&0.074&0.079 
        &0.072&0.073&0.078&0.086 
        &0.074&0.076&0.078&0.080\\
& PGSR~\cite{chen2024pgsr} 
        &0.185&0.185&0.189&0.192 
        &0.185&0.187&0.190&0.196
        &0.186&0.185&0.188&0.193
        &0.192&0.199&0.198&0.203 \\
& 2DGS~\cite{huang20242d} 
       &0.152&0.155&0.161&0.235
       &0.235&0.236&0.427&0.453 
       &0.196&0.195&0.200&0.201
       &0.201&0.193&0.199&0.202    
                \\
& MVGSR 
        &0.037 &0.038 &0.039 &0.044 
        &0.038 &0.041 &0.043 &0.047
        &0.045 &0.044 &0.047 &0.05
        &0.052 &0.054 &0.056 &0.058\\ \midrule 
        \toprule

\multicolumn{2}{l|}{scan} & \multicolumn{4}{c|}{114} & \multicolumn{4}{c|}{118} & \multicolumn{4}{c|}{122}& \multicolumn{4}{c}{Avg.} \\ \cline{1-18}

\multicolumn{2}{l|}{rate} & \multicolumn{1}{c|}{0.3} & \multicolumn{1}{c|}{0.5} & \multicolumn{1}{c|}{0.8} & 1.0 & \multicolumn{1}{c|}{0.3} & \multicolumn{1}{c|}{0.5} & \multicolumn{1}{c|}{0.8} & 1.0 & \multicolumn{1}{c|}{0.3} & \multicolumn{1}{c|}{0.5} & \multicolumn{1}{c|}{0.8} & 1.0& \multicolumn{1}{c|}{0.3} & \multicolumn{1}{c|}{0.5} & \multicolumn{1}{c|}{0.8} & 1.0 \\  \cline{1-18}
\multirow{3}{*}{PSNR$\uparrow$} 
& 3DGS~\cite{kerbl20233d} 
        &31.59&27.46&28.19&29.55 
        &35.13&34.89&32.51&32.60
        &32.69&31.54&29.48&27.09 
        &28.74&28.23&27.12&26.33\\
& PGSR~\cite{chen2024pgsr} 
        &32.47&32.54&32.02&31.61
        &37.23&37.04&35.71&35.67
        &36.71&36.01&35.67&34.46
        &32.97&32.44&31.97&31.30\\
& 2DGS~\cite{huang20242d} 
        &32.73&32.47&26.54&31.52
        &37.30&37.12&35.56&35.18
        &36.74&35.89&35.52&35.32
        &32.08&31.86&30.84&30.43\\
& MVGSR 
        &33.71 &33.49 &32.95 &32.67 
        &40.24 &40.13 &38.47 &38.75
        &40.82 &40.76 &39.27 &38.44
        &35.58 &35.09 &34.06 &32.98\\  \cline{1-18}
\multirow{3}{*}{SSIM$\uparrow$} 
& 3DGS~\cite{kerbl20233d} 
        &0.965&0.947&0.953&0.959 
        &0.968&0.966&0.960&0.962
        &0.972&0.967&0.958&0.948
        &0.956&0.953&0.948&0.943\\
& PGSR~\cite{chen2024pgsr}
        &0.924&0.925&0.922&0.923
        &0.933&0.933&0.928&0.930
        &0.931&0.928&0.927&0.925
        &0.928&0.926&0.924&0.921\\
& 2DGS~\cite{huang20242d} 
        &0.931&0.929&0.872&0.927 
        &0.935&0.934&0.930&0.929
        &0.931&0.930&0.930&0.927
        &0.918&0.927&0.918&0.900\\
& MVGSR  
        &0.97 &0.969 &0.967 &0.967 
        &0.982 &0.982 &0.978 &0.98
        &0.986 &0.986 &0.984 &0.982
        &0.970 &0.970 &0.967 &0.965\\  \cline{1-18}
\multirow{3}{*}{LPIPS$\downarrow$} 
& 3DGS~\cite{kerbl20233d} 
        &0.048&0.066&0.060&0.058
        &0.053&0.055&0.065&0.062
        &0.033&0.041&0.050&0.061
       &0.061&0.064&0.071&0.078\\
& PGSR~\cite{chen2024pgsr} 
        &0.170&0.170&0.176&0.176
        &0.194&0.192&0.205&0.199
        &0.196&0.210&0.206&0.202
        &0.158&0.162&0.165&0.168\\
& 2DGS~\cite{huang20242d} 
        &0.194&0.196&0.291&0.201
        &0.229&0.232&0.238&0.239
        &0.236&0.242&0.237&0.246
        &0.182&0.175&0.190&0.215\\
& MVGSR 
        &0.044 &0.045 &0.049 &0.05 
        &0.034 &0.035 &0.04 &0.038
        &0.022 &0.023 &0.026 &0.03
        &0.042 &0.044 &0.047 &0.051\\ \hline
\end{tabular}
}
\caption{Quantitative results of novel view synthesis (PSNR$\uparrow$) on DTU-Robust. }
\label{table_1_render_psnr}
\end{table*}

\section{More Results on  TnT-Robust Dataset}
\label{tntrobust}
\begin{figure*}[h]
    \centering
    \includegraphics[width=\linewidth]{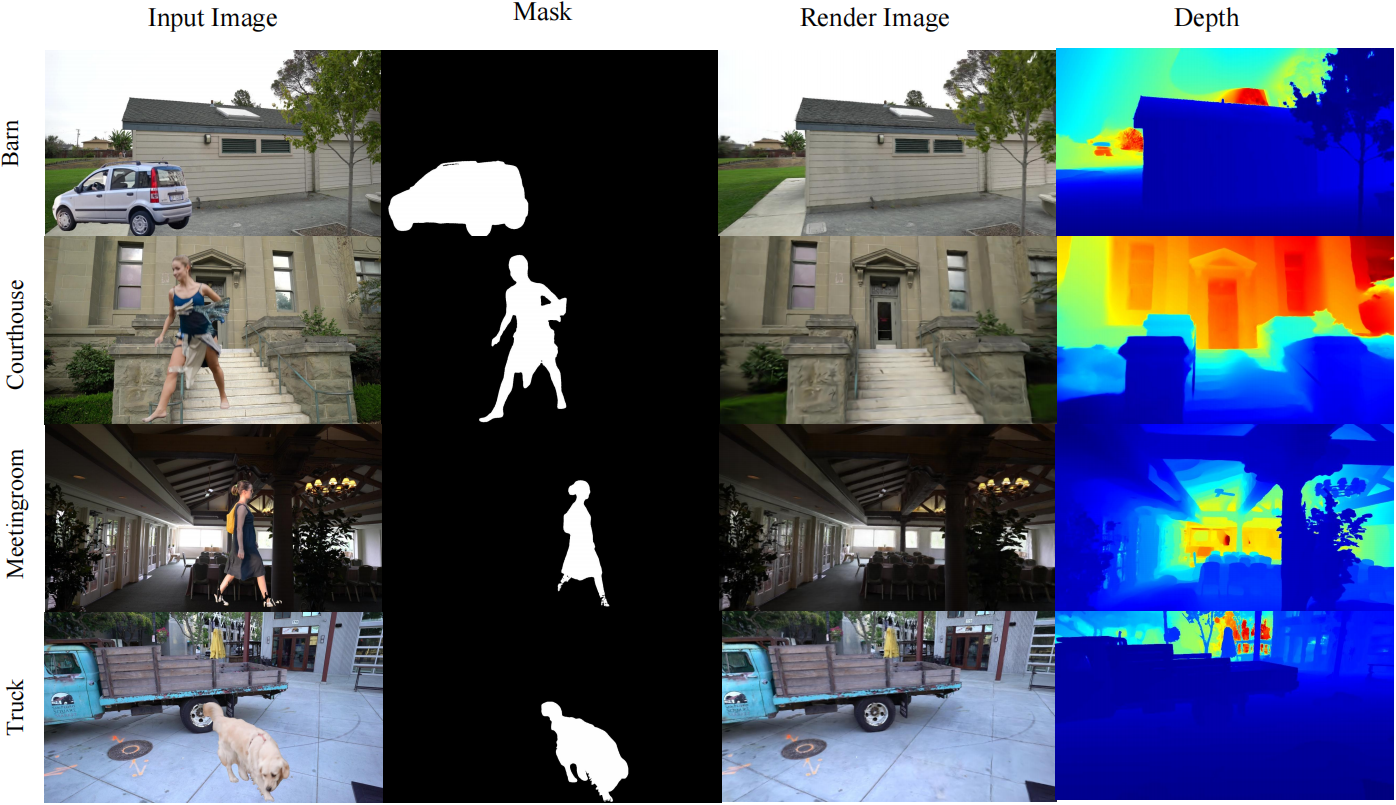}
    \caption{More results on TnT-Robust dataset}
    \label{fig:tnt-robust-more}
\end{figure*}

Tab. \ref{table_tnt2} and Fig. \ref{fig:tnt-robust-more} show more results on TnT-Robust Datset.


\section{More Results on  NeRF on-the-go Dataset}
\label{nerfonthego}
As shown in Figures \ref{fig:Mask} and \ref{fig:robustnerf2}, we present additional visualizations on the NeRF onthego and RobustNerf datasets, demonstrating clearer segmentation effects and artifact removal. Quantitative visualization metrics on the onthego dataset are further reported in Tab. \ref{on-the-go-recon}.

\end{document}